%% file: main.tex
\documentclass{article}

    \usepackage[preprint]{neurips_2025}

\usepackage[utf8]{inputenc} 
\usepackage[T1]{fontenc}    
\usepackage{hyperref}       
\usepackage{url}            
\usepackage{booktabs}       
\usepackage{amsfonts}       
\usepackage{nicefrac}       
\usepackage{microtype}      
\usepackage{xcolor}         

\usepackage{tcolorbox}
\usepackage{enumitem}
\usepackage{algorithmic}
\usepackage{algorithm}
\usepackage{multirow}%
\usepackage{mathrsfs}%
\usepackage[title]{appendix}%
\usepackage{textcomp}%
\usepackage{manyfoot}%
\usepackage{listings}%
\usepackage{anyfontsize}
\usepackage{color}
\usepackage{bbding}
\usepackage{longtable}
\usepackage{supertabular}
\usepackage{colortbl}
\usepackage{tocloft}
\usepackage{wrapfig} 
\usepackage{mathrsfs}
\usepackage{pifont} 
\usepackage{animate}
\usepackage{amsmath}
\usepackage{amssymb}
\usepackage{mathtools}
\usepackage{amsthm}

\usepackage[capitalize,noabbrev]{cleveref}

\theoremstyle{plain}
\newtheorem{theorem}{Theorem}[section]

\theoremstyle{definition}

\theoremstyle{remark}

\title{Revisiting Sparsity Constraint Under High-Rank Property in Partial Multi-Label Learning}

\author{Chongjie Si$^{*1}$, Yidan Cui$^{*1}$, Fuchao Yang$^2$, Xiaokang Yang$^1$, Wei Shen$^1$ \\
$^1$Shanghai Jiao Tong Univeristy, $^2$Southeast University\\
  \texttt{\{chongjiesi, wei.shen\}@sjtu.edu.cn}
}

\begin{document}

\maketitle

\input{sec/0.abstract}

\input{sec/1.intro}

\input{sec/2.related}

\input{sec/3.approach}

\input{sec/4.solution}

\input{sec/5.experiments}

\input{sec/6.conclusion}

\bibliography{main}
\bibliographystyle{plain}


\newpage

\input{sec/appendix}


\end{document}

%% file: sec/0.abstract.tex
\begin{abstract}
Partial Multi-Label Learning (PML) extends the multi-label learning paradigm to scenarios where each sample is associated with a candidate label set containing both ground-truth labels and noisy labels. 
Existing PML methods commonly rely on two assumptions: sparsity of the noise label matrix and low-rankness of the ground-truth label matrix. 
However, these assumptions are inherently conflicting and impractical for real-world scenarios, where the true label matrix is typically full-rank or close to full-rank. 
To address these limitations, we demonstrate that the sparsity constraint contributes to the high-rank property of the predicted label matrix. 
Based on this, we propose a novel method Schirn, which introduces a sparsity constraint on the noise label matrix while enforcing a high-rank property on the predicted label matrix. 
Extensive experiments demonstrate the superior performance of Schirn compared to state-of-the-art methods, validating its effectiveness in tackling real-world PML challenges.

\end{abstract}

%% file: sec/1.intro.tex
\section{Introduction}

\begin{wrapfigure}{r}{0.45\textwidth}
\includegraphics[width=0.05\linewidth, bb=0 0 37 200]{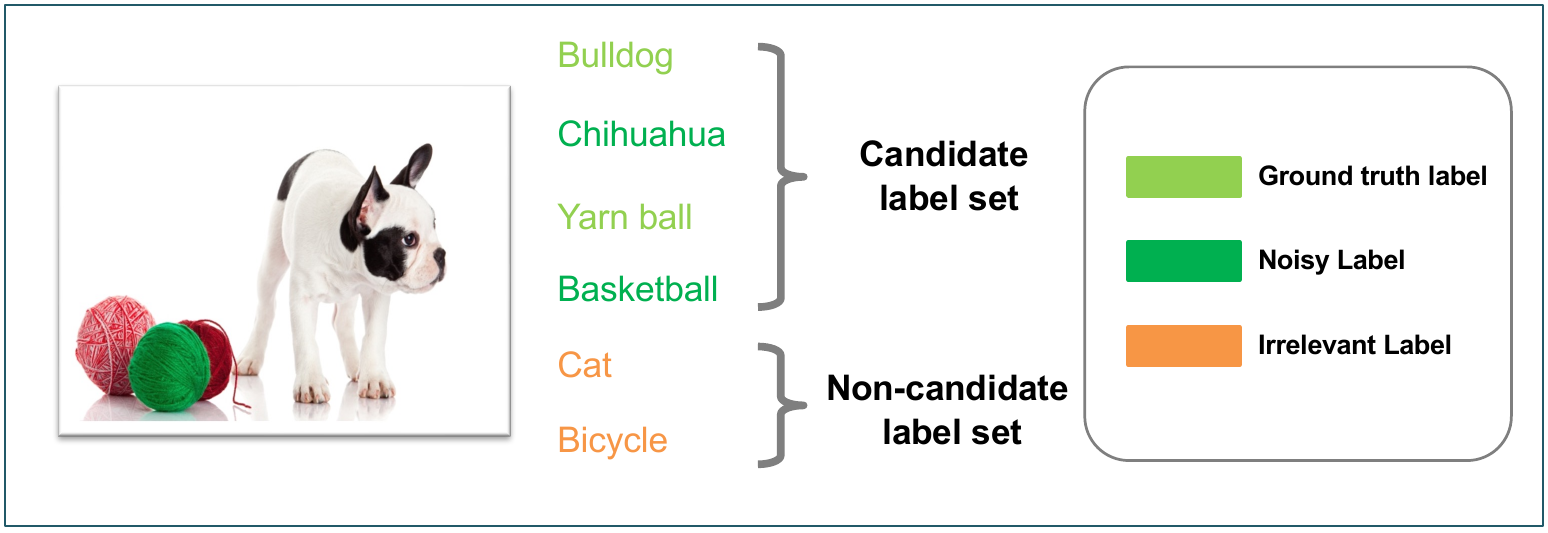}
  \caption{An example of partial multi-label learning.}
  \label{fig:frame}
\end{wrapfigure}

Multi-label learning (MLL) \cite{zhang2013review, zhang2014lift, zhang2018binary, si2023multi} is a fundamental machine learning paradigm where each instance is associated with multiple labels. 
It has been widely applied in diverse domains such as image annotation \cite{kuznetsova2020open, qi2007correlative}, bioinformatics \cite{cheng2017iatc, zhang2006multilabel}, web mining \cite{tang2009large}, and information retrieval \cite{lai2016instance} due to its ability to model complex relationships between instances and their associated labels.
However, in practical scenarios, obtaining fully annotated data with precise labels is often infeasible due to the high cost of labeling and the inherent ambiguity of many tasks. 
Instead, annotators typically provide a set of candidate labels, which includes both ground-truth and noisy labels, as shown in Fig. \ref{fig:frame}. 
This motivates a new learning paradigm known as Partial Multi-Label Learning (PML) \cite{xie2018partial,sun2019partial,yu2020partial,yang2024noisy}.

Formally, let $\mathcal{X} = \mathbb{R}^d$ represent the $d$-dimensional feature space, and let $\mathcal{Y} = \{1, 2, \dots, l\}$ denote the label space with $l$ possible labels. 
Consider a partial multi-label training set $\mathcal{D} = \{(\mathbf{x}_i, \mathcal{C}_i) \,|\, 1 \leq i \leq n\}$, where $n$ is the number of training samples. 
Here, $\mathbf{x}_i \in \mathcal{X}$ is the $d$-dimensional feature vector for the $i$-th instance, and $\mathcal{C}_i \subseteq \mathcal{Y}$ is its candidate label set. 
The goal of PML is to learn a multi-label classifier $f: \mathcal{X} \to 2^{\mathcal{Y}}$ from $\mathcal{D}$, which can accurately predict the true labels for unseen instances. 
Beyond the inherent challenges of MLL, i.e., the exponentially large number of possible label combinations, PML introduces an additional layer of complexity: the true labels are not directly accessible during training.
This lack of explicit ground-truth labels requires the model to disambiguate between correct and noisy labels within the candidate label set, significantly increasing the difficulty of the learning task.

To address the challenges posed by PML, recent years have witnessed the development of numerous innovative techniques \cite{sun2021partial, wang2022partial, zhao2022partial, sun2021global, hang2023partial}, particularly those rely on the \textbf{sparse assumption} of noisy labels and the \textbf{low-rank assumption} of ground-truth labels \cite{yang2024noisy, sun2019partial, xie2021partial, sun2021partial}. 
These approaches assume that the noise label matrix is sparse and aim to optimize its $\ell_0$-norm, as the sparsity constraint helps reduce the model’s generalization error \cite{yang2024noisy}. 
Simultaneously, they leverage the observation from MLL that ground-truth labels are often correlated, imposing low-rank constraints on the classifier’s weights to constrain the low-rank property of the predicted label matrix. 
By jointly optimizing these assumptions within a unified loss function, these approaches aim to construct models that can effectively disambiguate noisy labels and exploit label correlations. 
However, is this kind of approach truly reasonable?

\textbf{We first emphasize that the assumptions of sparsity in the noise label matrix and the low-rank property of the true label matrix are inherently conflicting.} 
When the noise label matrix is sparse, the predicted label matrix can be treated as a minor perturbation of the observed label matrix. 
According to Wedin’s theorem \cite{wedin1972perturbation}, sparse perturbations have a minimal effect on the singular values of the original matrix, meaning that the rank of the predicted label matrix should remain closely aligned with the rank of the observed matrix.
If the predicted label matrix is assumed to be low-rank, the observed matrix would also need to exhibit low-rank properties under sparse noise. 
However, as shown in Table \ref{tab:high-rank}, the observed label matrices across real-world PML datasets are consistently full rank.
This observation strongly challenges the compatibility of these assumptions, indicating that \textbf{if the true label matrix were to maintain a low-rank structure, the noise label matrix could not plausibly be sparse.}

\begin{table}[ht!]
    \centering
    \renewcommand{\arraystretch}{1}
    \caption{The rank of the label matrix across commonly used real-world PML and extremely large MLL datasets. Here, $n$ denotes the number of samples, $l$ denotes the number of labels, $r(\mathbf{Y})$ and $r(\mathbf{Y}_g)$ represent the rank of the observed label matrix and true label matrix, respectively. ``Ratio'' refers to the sparsity level of the noise label matrix when the rank of $\mathbf{Y}$ is reduced. We randomly set one column of the observed matrix to zero and calculate the proportion of elements that are changed.}
    \resizebox{\textwidth}{!}{
    \begin{tabular}{c c c c c c | c c c c}
    \toprule
        PML Dataset & $n$ & $l$ & $r(\mathbf{Y})$ & $r(\mathbf{Y}_g)$ & Ratio & PML Dataset & $n$ & $l$ & $r(\mathbf{Y})$ ($r(\mathbf{Y}_g)$) \\ 
        \midrule
         Music\_emotion & 6833 & 11 & 11 & 11 & 2.70\% & LF-Amazon-131k & 294805 & 131073 & 126051 \\
         Music\_style & 6839  & 10 & 10 & 10 & 5.36\% & see-also & 693082 & 312330 & 271563 \\
         YeastMF & 1408 & 39 & 39 & 36 & 0.22\% & wiki-titles & 1813391 & 501070 & 495540 \\
         YeastCC & 1771 & 50 & 50 & 45 & 0.28\% & Amazon-titles & 2248619 & 1305265 & 1291202 \\
         YeastBP & 3794 & 217 & 217 & 200 & 0.03\% & Amazon-cat & 1186239 & 13330 & 13307  \\
        \bottomrule
    \end{tabular}}
    \label{tab:high-rank}
\end{table}

Furthermore, we argue that \textbf{the true label matrix should be full-rank or close to full-rank}. 
While existing methods assume low-rankness due to correlations between labels, such correlations do not eliminate label independence. 
In practice, connections between labels do not guarantee co-occurrence across all samples, and thus, they cannot reduce the rank of the label matrix.
As shown in Table \ref{tab:high-rank}, the rank of the true label matrix is consistently full-rank or near full-rank across real-world datasets. 
This observation is further supported by prior studies \cite{si2023multi, ma2021expand}, which have similarly noted that true label matrices in MLL scenarios typically exhibit full-rank properties.

Given these findings, \textbf{we propose a reevaluation of the relationship between sparsity and the rank characteristics of the label matrix.} Specifically, the sparsity of the noise label matrix should not be viewed as a means of enforcing low-rankness in the true label matrix but rather as a mechanism for preserving the high-rank structure of the predicted label matrix.
First, based on Wedin’s theorem, when the noise label matrix is sparse, the rank of the predicted label matrix closely matches that of the observed label matrix, as sparse noise minimally impacts rank properties. Second, as illustrated in Table \ref{tab:high-rank}, when the noise label matrix maintains sufficient sparsity (below a certain threshold), the rank of the predicted label matrix remains same as that of the observed matrix. This strongly suggests that sparsity primarily functions to ensure the high-rank or full-rank structure of the predicted label matrix, which is critical for capturing the complexity and richness of real-world multi-label data.

Therefore, in this paper, we propose a novel method named \textbf{Schirn} (\textbf{S}parsity \textbf{c}onstraint under \textbf{hi}gh-\textbf{r}a\textbf{n}k property). Specifically, our approach enforces a sparsity constraint on the noise label matrix while simultaneously imposing a high-rank constraint on the predicted label matrix. This dual constraint ensures that the model effectively separates noisy labels while maintaining the richness and complexity of the label structure. 
By addressing the inherent limitations of existing assumptions, Schirn provides a robust and principled framework for tackling real-world PML problems.
Extensive experiments conducted on five real-world partial multi-label datasets and six synthetic datasets validate the effectiveness of the proposed Schirn method, highlighting its ability to handle PML.

%% file: sec/2.related.tex
\section{Related Work}\label{sec related}


\subsection{Multi-Label Learning (MLL)}
MLL addresses the problem of associating each sample with multiple labels, with a primary focus on uncovering and utilizing label correlations. 
Depending on the type of correlations considered, MLL methods can be categorized into three main groups: first-order label correlations, second-order label correlations, and high-order label correlations.
First-order methods treat each label independently, ignoring interactions between labels \cite{BOUTELL2004Learning, zhang2007mlds}. 
Second-order methods model pairwise correlations between labels, leveraging mutual dependencies between pairs of classes \cite{Johannes2008Multilabel, zhang2014lift, huang2015learning}. 
High-order approaches capture the complex dependencies among multiple labels simultaneously, considering the collective influence of all labels \cite{2009Classifier1212,tsoumakas2010random,2017Multisdfsaf,si2023multi,9084698}.
Despite their effectiveness in fully annotated settings, traditional MLL methods are not directly applicable to PML, as they assume access to precise ground-truth labels, which is not the case in PML. The presence of noisy labels in candidate sets further complicates the application of existing MLL techniques to PML scenarios.

\subsection{Partial Multi-label Learning (PML)}
PML focuses on resolving label ambiguity by identifying the ground-truth labels from the candidate label set. Existing PML approaches can be broadly divided into two categories: two-stage methods and end-to-end methods.

Two-stage methods divide the process into separate phases.
The first stage focuses on label disambiguation, often employing techniques like label propagation to extract more reliable labels from the candidate set. 
The second stage uses the refined labels as ground-truth labels to train a multi-label classifier.
For example, \cite{zhang2020partial} use maximum a posterior reasoning or virtual label splitting, while \cite{wang2019discriminative} employ gradient-boosting models to achieve classification.
End-to-end methods iteratively perform label disambiguation and model training within a unified framework. 
For instance, \cite{wang2023deep, xu2020partial} incorporate sample similarity in the feature space and label correlations simultaneously.
These methods alternate between refining noisy labels and updating the classifier to achieve label disambiguation.

Recently, several works \cite{sun2019partial, xie2021partial, sun2021partial} have leveraged the low-rank assumption for ground-truth labels and the sparse assumption for noisy labels to refine candidate sets while capturing label correlations. 
However, as discussed in the introduction, there is an inherent conflict between the sparsity of the noise label matrix and the low-rank property of the true label matrix. 
Furthermore, true label matrix in MLL is generally full-rank or close to full-rank.
Therefore, based on these findings, we re-explored the relationship between sparsity and rank properties and propose a new approach to address these issues.
In the following sections, we will introduce our proposed method in detail, which imposes a sparsity constraint on the noise label matrix under the assumption of high-rank properties for the true label matrix. 

%% file: sec/3.approach.tex
\section{Proposed Approach}
In this section, we first define the necessary notations and then provide a detailed introduction to Schirn. 
Schirn is designed to jointly constrain the sparsity of the noise label matrix and the high-rank property of the predicted label matrix, enabling more effective label disambiguation.

\subsection{Notation}

Denote $\mathbf{X} = [\mathbf{x}_1,\mathbf{x}_2,\cdots, \mathbf{x}_n]\in\mathbb{R}^{n\times d}$ is the instance matrix and $\mathbf{Y} = [\mathbf{y}_1, \mathbf{y}_2,\cdots,\mathbf{y}_n] \in \{0,1\}^{n\times l}$ is the observed label matrix, i.e., the candidate label matrix.
Here, $\mathbf{y}_i = [y_{i1}, y_{i1}, \cdots, y_{il}]$ is the label vector for sample $\mathbf{x}_i$, where $y_{ij}=1$ if the $j$-th label is one of the candidate labels of sample $\mathbf{x}_i$ and $y_{ij}=0$ otherwise.

\subsection{Ordinary Classifier}
Schirn begins by leveraging a weight matrix $\mathbf{W}\in\mathbb{R}^{d\times l}$ to map the instance matrix $\mathbf{X}$ to the predicted true label matrix $\mathbf{Y} - \mathbf{N}$ , where $\mathbf{N}\in\mathbb{R}^{n\times l}$ represents the noise label matrix. 
Typically, the noise label matrix $\mathbf{N}$ is subject to two constraints to ensure meaningful label disambiguation:
(1) $\mathbf{N} \in \{0, 1\}^{n \times l}$, which restricts all elements in $\mathbf{N}$ to binary values (0 or 1). 
Specifically, $N_{ij} = 1$ indicates that the $j$-th label of the sample $\mathbf{x}_i$ is a noisy label, while $N_{ij} = 0$ implies it is not. (2) For $\forall i, j$, $N_{ij} \leq Y_{ij}$.
This constraint ensures that noisy labels can only exist within the candidate label set, i.e., they cannot appear in non-candidate labels.

To construct the ordinary classifier, we use the following least-squares loss function as the foundation of our model:
\begin{equation}
    \begin{aligned}
    & \min_{\mathbf{W}, \mathbf{N}} \|\mathbf{X}\mathbf{W} - (\mathbf{Y} - \mathbf{N}) \|_F^2 + \lambda \|\mathbf{W}\|_F^2 \\
    & {\rm s.t.}\quad \mathbf{N} \in\{0,1\}^{n\times l}, \forall i,j, N_{ij}\leq Y_{ij}.
    \end{aligned}
    \label{eq ordinary classifier}
\end{equation}
Here, $\|\cdot\|_F$ represents the Frobenius norm of a matrix, and minimizing the regularization term $\|\mathbf{W}\|_F^2$ will control the complexity of the weight matrix $\mathbf{W}$, helping to prevent overfitting. 
The hyper-parameter $\lambda>0$ balances the trade-off between fitting the data and controlling model complexity.

\subsection{Sparsity Constraint Under High-Rank Property}
Sparsity on the noise label matrix is a widely used technique that helps reduce the model’s generalization error \cite{yang2024noisy, sun2019partial, xie2021partial, sun2021partial}.
Additionally, considering the well-known label correlations in multi-label learning, existing methods often impose a low-rank constraint on the weight matrix $\mathbf{W}$ alongside sparsity, which aims to maintain the low-rank property of the predicted label matrix while effectively capturing label correlations.
However, as highlighted in the introduction, sparsity and the low-rank property of the predicted label matrix are inherently conflicting, and the true label matrix in MLL is typically full-rank or close to full-rank. 
Therefore, a more reasonable approach is to constrain the sparsity of the noise label matrix while preserving the high-rank property of the predicted label matrix. 
We here first provide theoretical proof of the relationship between the sparsity and high-rank property.
\begin{theorem}
Let $\mathbf{Y} \in \mathbb{R}^{n \times l}$ be a full-rank matrix with $\text{rank}(\mathbf{Y}) = \min(n, l)$, and let $\mathbf{N} \in \mathbb{R}^{n \times l}$ be a sparse binary matrix satisfying $\|\mathbf{N}\|_0 \leq \epsilon$, where $\|\cdot\|_0$ represents the $\ell_0$-norm and $\epsilon$ is a small positive integer smaller than $n$ and $l$.
Then, the rank of the matrix $\mathbf{Y}_g = \mathbf{Y} - \mathbf{N}$ satisfies:
\begin{equation}
    \text{rank}(\mathbf{Y}_g) \geq \min(n, l) - \Delta,
\end{equation}
where $\Delta$ is a very small positive integer that depends on the sparsity level $\epsilon$ of $\mathbf{N}$.
\label{theorem}
\end{theorem}
The detailed proofs of Theorem \ref{theorem} are shown in Appendix \ref{sec appendix proof}.
Theorem \ref{theorem} demonstrates that the rank of a full-rank matrix remains high under sparse perturbations, validating the feasibility of simultaneously imposing a sparsity constraint on the noise label matrix and maintaining the high-rank structure of the predicted label matrix. 
This balance ensures that the model retains the complexity and richness of label structures while effectively handling noisy labels.

Specifically, the sparsity constraint on the noise label matrix under the high-rank property of the predicted label matrix can be formulated as:
\begin{equation}
    \min_{\mathbf{W}, \mathbf{N}} \alpha \|\mathbf{N}\|_0 - \beta\text{rank}(\mathbf{X}\mathbf{W}),
    \label{eq sparse high rank}
\end{equation}
where $\alpha>0$ is a hyper-parameter controlling the emphasis on the sparsity of the noise matrix $\mathbf{N}$, while  $\beta>0$ is another hyper-parameter determines the importance of maximizing the rank of  $\mathbf{X}\mathbf{W}$.

Directly optimizing this objective is computationally intractable due to the non-convexity of both the $\ell_0$-norm and the rank function.
To simplify the optimization, recent advancements in matrix optimization suggest replacing the rank function with its convex surrogate, the nuclear norm, which is defined as the sum of the singular values of the matrix \cite{candes2012exact}. 
Additionally, since the noise label matrix $\mathbf{N}$ is a binary matrix, the $\ell_0$-norm can be replaced by the  $\ell_1$-norm, which sums the absolute values of matrix elements \cite{candes2005decoding}.
Incorporating these relaxations, the optimization problem in Eq. (\ref{eq sparse high rank}) can be rewritten as:
\begin{equation}
    \min_{\mathbf{W}, \mathbf{N}} \alpha \|\mathbf{N}\|_1 - \beta \|\mathbf{X}\mathbf{W}\|_*,
    \label{eq relax sparse}
\end{equation}
where $\|\cdot\|_1$ is the $\ell_1$ norm of a matrix, and $\|\cdot\|_*$ denotes the nuclear norm of a matrix.

\subsection{Overall Formulation}
By combining Eq. (\ref{eq ordinary classifier}) and Eq. (\ref{eq relax sparse}), the final objective function can be formulated as follows:
\begin{equation}
    \begin{aligned}
     & \min_{\mathbf{W}, \mathbf{N}} \quad \|\mathbf{X}\mathbf{W} - (\mathbf{Y} - \mathbf{N}) \|_F^2 + \alpha \|\mathbf{N}\|_1 - \beta \|\mathbf{X}\mathbf{W}\|_* + \lambda \|\mathbf{W}\|_F^2 \\
   & {\rm s.t.}\quad \mathbf{N} \in\{0,1\}^{n\times l}, \forall i,j, N_{ij}\leq Y_{ij}.
    \end{aligned}
    \label{eq final formulation}
\end{equation}
The proposed Schirn method effectively enforces sparsity on the noise label matrix while maintaining the high-rank property of the predicted label matrix, allowing the model to disambiguate noisy labels while preserving the richness of the label structure.
To the best of our knowledge, this is the first work to systematically investigate the relationship between sparsity and high-rank properties in partial multi-label learning, providing a novel perspective and solution to address the challenges of PML.

%% file: sec/4.solution.tex
\section{Optimization}

In Eq. (\ref{eq final formulation}), we have two variables, $\mathbf{W}$ and $\mathbf{N}$, that need to be optimized. We first reformulate the equation as follows:
\begin{align}
     & \min_{\mathbf{W}, \mathbf{N},\mathbf{C}} \|\mathbf{C} - (\mathbf{Y} - \mathbf{N}) \|_F^2 + \alpha \|\mathbf{N}\|_1  - \beta \|\mathbf{C}\|_* + \lambda \|\mathbf{W}\|_F^2 \nonumber\\
   & {\rm s.t.}\quad \mathbf{C} = \mathbf{XW}, \mathbf{N} \in\{0,1\}^{n\times l}, \forall i,j, N_{ij}\leq Y_{ij}.
   \label{eq convert formulation}
\end{align}
The optimization problem in Eq. (\ref{eq convert formulation}) can be solved using the Augmented Lagrange Multiplier (ALM) method \cite{lin2010augmented,rockafellar1974augmented}.
The ALM framework introduces a dual variable to handle the equality constraint $\mathbf{C} = \mathbf{XW}$ and minimizes the following augmented Lagrange function:
\begin{align}
    & \min_{\mathbf{W}, \mathbf{N}, \mathbf{C}, \mathbf{\Lambda}}  \|\mathbf{C} - (\mathbf{Y} - \mathbf{N}) \|_F^2 + \alpha \|\mathbf{N}\|_1 - \beta \|\mathbf{C}\|_*  + \lambda \|\mathbf{W}\|_F^2 
    + \langle \mathbf{\Lambda}, \mathbf{XW} - \mathbf{C} \rangle + \frac{\mu}{2} \|\mathbf{XW} - \mathbf{C}\|_F^2. \nonumber\\
    & {\rm s.t.} \quad \mathbf{N} \in\{0,1\}^{n\times l}, \forall i,j, N_{ij}\leq Y_{ij},
    \label{eq admm}
\end{align}
where $\mathbf{\Lambda}\in\mathbb{R}^{n\times l}$ is the Lagrange multiplier and $\mu$ is the penalty parameter.
We then can optimize Eq. (\ref{eq admm}) by solving the following subproblems alternatively and iteratively.

\subsection{$\mathbf{W}$ Subproblem}
With other variables fixed, the $\mathbf{W}$ subproblem becomes
\begin{equation}
    \min_{\mathbf{W}}  \lambda \|\mathbf{W}\|_F^2 
    + \langle \mathbf{\Lambda}, \mathbf{XW} - \mathbf{C} \rangle + \frac{\mu}{2} \|\mathbf{XW} - \mathbf{C}\|_F^2.
    \label{eq W subproblem}
\end{equation}
The closed-form solution in Eq. (\ref{eq W subproblem}) is obtained by setting the first-order derivative of the objective function with respect to  $\mathbf{W}$ to zero, leading to
\begin{equation}
\mathbf{W} = (\mu \mathbf{X}^T \mathbf{X} + 2\lambda \mathbf{I}_{d\times d})^{-1} (\mu \mathbf{X}^T \mathbf{C} - \mathbf{X}^T \mathbf{\Lambda}),
\label{eq W solution}
\end{equation}
where $\mathbf{I}_{d\times d} \in \mathbb{R}^{d\times d}$ is an identity matrix.

\subsection{$\mathbf{N}$ Subproblem}
With other variables fixed, the $\mathbf{N}$ subproblem becomes
\begin{equation}
    \begin{aligned}
    & \min_{\mathbf{N}} f(\mathbf{N})  + \alpha g(\mathbf{N}) \\
    & {\rm s.t.} \quad \mathbf{N} \in\{0,1\}^{n\times l}, \forall i,j, N_{ij}\leq Y_{ij},
    \label{eq N subproblem}
\end{aligned}
\end{equation}
where $f(\mathbf{N}) = \|\mathbf{N} - (\mathbf{Y} - \mathbf{C}) \|_F^2 $ and $g(\mathbf{N}) = \|\mathbf{N}\|_1$ are both convex functions. 
We first verify that $f(\mathbf{N})$ has a Lipschitz continuous gradient $\nabla f(\mathbf{N})$ \cite{beck2009fast}, which is calculated as $\nabla f(\mathbf{N}) = 2(\mathbf{N} - (\mathbf{Y}-\mathbf{C}))$. 
For two different $\mathbf{N}_1$ and $\mathbf{N}_2$, we have
\begin{equation}
    \|\nabla f(\mathbf{N}_1) - \nabla f(\mathbf{N}_2)\|_F = 2\|\mathbf{N}_1-\mathbf{N}_2\|_F \leq L_f\|\Delta\mathbf{N}\|_F,
\end{equation}
where $\Delta\mathbf{N} = \mathbf{N}_1-\mathbf{N}_2$  and $L_f$ is the Lipschitz constant. 
From this, we observe that $f(\mathbf{N})$ is Lipschitz continuous with $L_f = 2$.
Given that $f(\mathbf{N})$ satisfies the required properties and the optimization problem in Eq. (\ref{eq N subproblem}) adheres to the form suitable for the iterative shrinkage thresholding algorithm (ISTA) \cite{beck2009fast}, we can employ ISTA to solve this problem efficiently, with the optimization process as
\begin{equation}
    \begin{aligned}
    \mathbf{N} = &\arg\min_{\mathbf{N}} \frac{L_f}{2} \|\mathbf{N} - \mathbf{M} \|_F^2 + \alpha \|\mathbf{N}\|_1
    = \arg\min_{\mathbf{N}} \frac{1}{2} \|\mathbf{N} - \mathbf{M} \|_F^2 + \frac{\alpha}{L_f} \|\mathbf{N}\|_1  \\
     {\rm s.t.} \quad &  \mathbf{N} \in\{0,1\}^{n\times l}, \forall i,j, N_{ij}\leq Y_{ij},
    \label{eq N optimization}
\end{aligned}
\end{equation}
where $\mathbf{M} = \mathbf{N} - \frac{1}{L_f}\nabla f(\mathbf{N}) = \mathbf{Y} - \mathbf{C}$ with $L_f = 2$.
The solution of Eq. (\ref{eq N optimization}) is 
\begin{equation}
    \mathbf{N} = \mathcal{T}_{\mathbf{Y}}\left( \mathcal{T}_{\rm sgn} \left( \mathcal{S}_{\alpha/L_f} \left( \mathbf{M}\right) \right) \right).
    \label{eq N solution}
\end{equation}
Here, $\mathcal{T}_{\mathbf{Y}}$ is the element-wise thresholding operator that $\mathcal{T}_\mathbf{Y}(\mathbf{A}) = [\min({A}_{ij}, {Y}_{ij})]_{ij}$ for $\mathbf{A}$ being a matrix.
$\mathcal{T}_{\rm sgn}(a)$ is the Sign function that returns 1 if $a > 0$ and 0 otherwise.
$ \mathcal{S}_{\alpha/L_f}$ is the shrinkage operator \cite{zhuang2012non} that
\begin{equation}
    \mathcal{S}_{\epsilon}(a) =
\begin{cases}
a - \epsilon, & \text{if } a > \epsilon, \\
a + \epsilon, & \text{if } a < -\epsilon, \\
0, & \text{otherwise.}
\end{cases}
\end{equation}

\subsection{$\mathbf{C}$ Subproblem}

By fixing other variables, the $\mathbf{C}$ subproblem can be expressed as
\begin{equation}
    \min_{\mathbf{C}}\quad  (1+\frac{\mu}{2})\|\mathbf{C} - \mathbf{G} \|_F^2  - \beta \|\mathbf{C}\|_*,
    \label{eq C subproblem}
\end{equation}
where $\mathbf{G} = \frac{2\mathbf{Y} - 2\mathbf{N} +\mathbf{\Lambda} + \mu \mathbf{X}\mathbf{W}}{2+\mu}$.
We here use singular value shrinkage theorem \cite{cai2010singular} to solve this optimization problem.
We decompose matrix $\mathbf{G}$ using singular value decomposition (SVD) as $\mathbf{G} = \mathbf{U}\mathbf{\Sigma}\mathbf{V}^\mathsf{T}$. 
The solution of Eq. (\ref{eq C subproblem}) is 
\begin{equation}
    \mathbf{C} = \mathbf{U} \max \left(0, \mathbf{\Sigma}+\frac{2\beta}{2+\mu} \mathbf{I} \right)  \mathbf{V}^\mathsf{T}.
    \label{eq C solution}
\end{equation}

\subsection{Lagrange Term Subproblem}
Finally, the Lagrange multiple matrix $\mathbf{\Lambda}$ and penalty parameter $\mu$ are updated based on the following formulations:
\begin{align}
     & \mathbf{\Lambda} \leftarrow \mathbf{\Lambda}  + \mu (\mathbf{X}\mathbf{W} - \mathbf{C})  \label{eq Lambda solution}\\
    & \mu \leftarrow \min(\mu_{\max}, \rho \mu), \label{eq mu solution}
\end{align}
where $\rho$ is a positive scalar typically set as $\rho=1.1$, and $\mu_{\max}$ is the upper bound for $\mu$, which is set as $\mu=10$. 
The pseudo code of Schirn is shown in Algorithm \ref{alg:schirn} in the Appendix.

%% file: sec/5.experiments.tex
\section{Experiments}

\subsection{Dataset}
We evaluated the performance of Schirn on 11 datasets, consisting of 5 real-world partial multi-label datasets\footnote{\url{http://palm.seu.edu.cn/zhangml/}} and 6 synthetic datasets\footnote{\url{http://www.uco.es/kdis/mllresources/}}.
For the real-world datasets YeastMF, YeastCC, and YeastBP, we observed that a significant portion of samples lacked ground-truth labels and contained only a small number of noisy labels. 
To ensure the integrity and fairness of the experiments, we removed these samples without ground-truth labels. 
For the remaining samples, we introduced additional noise by randomly adding $r$ noisy labels, where $r$ was set to 3, 5, and 10 for the datasets YeastMF, YeastCC, and YeastBP, respectively.
For the synthetic datasets, they are derived from widely used multi-label datasets by introducing controlled noise, following the procedure described in \cite{yang2024noisy,lyu2020partial,lyu2021prior}.
Specifically, for each sample, we randomly selected $r$ negative labels and added them as noisy labels to the candidate label set. 
As a result, the candidate label set for the synthetic datasets comprises all ground-truth labels along with $r$ additional noisy labels.
The selection of $r$ was determined based on the number of classes in each dataset. For datasets with fewer than 10 classes, we set $r \in \{1, 2, 3\}$; For datasets with between 10 and 100 classes, we set $r \in \{3, 7, 11\}$; For datasets with over 100 classes, we set $r \in \{10, 20, 30\}$.
For more details on the datasets, please refer to Table. \ref{tab:dataset_details}.

\subsection{Compared Approaches and Metrics}
We compared our proposed method against 9 state-of-the-art PML approaches, including NLR \cite{yang2024noisy}, FPML \cite{yu2018feature}, PML-LRS \cite{sun2019partial}, PML-NI \cite{xie2021partial}, PARTICLE-MAP and PARTICLE-VLS (P-MAP and P-VLS for short) \cite{zhang2020partial}, PAKS \cite{lyu2021prior}, GLC \cite{sun2021global}, and PARD \cite{hang2023partial}.
Specifically, NLR directly impose a sparsity constraint on the noise label matrix to suppress irrelevant labels.
In addition to sparsity constraints, PML-LRS and PML-NI further introduce low-rank regularization on the weight matrix.
Other methods utilize label propagation, probabilistic reasoning, or global label correlations to disambiguate noisy labels and refine the candidate label set iteratively.
Each method was implemented with its recommended hyper-parameter settings as reported in the respective literature to ensure a fair comparison.

For evaluation, we adopted five widely-used multi-label performance metrics \cite{zhang2006multilabel,gibaja2015tutorial,si2023multi}: \textit{average precision, ranking loss, coverage, hamming loss, and one-error}.
To ensure the reliability of the results, we performed five-fold cross-validation on each dataset. 
For each method, we recorded the mean value and standard deviation for all evaluation metrics.

\subsection{Hyper-parameter Settings}
Schirn has three hyper-parameters $\alpha$, $\beta$ and $\lambda$, which control the sparsity regularization, high-rank term and the model complexity, respectively.
$\alpha$ was selected from $[0.1, 2]$ with step size 0.1, $\beta$ was tuned from $[0.01, 0.1]$ with step size 0.01, and $\lambda$ was tuned from $\{0.1,10,100,250,1000\}$.
The parameters for optimizing Lagrange term is set as follows: $\mu_{\max} = 10$ and $\rho=1.1$. The total number of iterations was set to 100.

\subsection{Experimental Results}

The experimental results in Tables. \ref{tab:average precision}-\ref{tab:ranking loss} and \ref{tab:coverage}-\ref{tab:one error} in the Appendix clearly demonstrate that Schirn consistently outperforms state-of-the-art methods across multiple evaluation metrics.
Notably, on the YeastMF dataset, Schirn improves the average precision from 0.510 to 0.544 and reduces the ranking loss from 0.258 to 0.197, showcasing its ability to achieve both better label disambiguation and more accurate predictions. 
This trend is consistent across other datasets as well, where Schirn demonstrates its superiority in capturing label correlations while addressing noisy labels effectively.

\begin{table*}[!ht]
\renewcommand\arraystretch{0.9} 
\centering
\caption{Experimental results on \textit{average precision} (\%, higher is better $\uparrow$). $\bullet$ denotes the best result among all methods.}
\resizebox{\textwidth}{!}{
\begin{tabular}{c c l l l l l l l l l l}
    \toprule
    
    Data & r & \multicolumn{1}{c}{Schirn} & \multicolumn{1}{c}{NLR} & \multicolumn{1}{c}{FPML} & \multicolumn{1}{c}{PML-LRS} & \multicolumn{1}{c}{PML-NI} & \multicolumn{1}{c}{P-MAP} & \multicolumn{1}{c}{P-VLS} & \multicolumn{1}{c}{PAKS} & \multicolumn{1}{c}{GLC} & \multicolumn{1}{c}{PARD} \\

    \midrule
     
    Music\_emotion & & 62.6 $\pm$ 0.4 $\bullet$ & 58.6 $\pm$ 1.1 & 55.6 $\pm$ 0.7 & 61.6 $\pm$ 1.0 & 60.8 $\pm$ 1.1 & 58.9 $\pm$ 2.0 & 61.1 $\pm$ 1.2 & 61.3 $\pm$ 1.2 & 61.5 $\pm$ 1.1 & 60.8 $\pm$ 1.1 \\

    Music\_style & & 75.0 $\pm$ 0.3 $\bullet$ & 71.4 $\pm$ 0.5 & 68.9 $\pm$ 0.5 & 73.3 $\pm$ 0.8 & 73.8 $\pm$ 0.8 & 72.2 $\pm$ 0.5 & 71.7 $\pm$ 0.7 & 72.8 $\pm$ 0.6 & 73.6 $\pm$ 1.0 & 73.2 $\pm$ 1.1 \\

    YeastMF & & 54.4 $\pm$ 1.1 $\bullet$ & 51.0 $\pm$ 1.6 & 33.8 $\pm$ 2.6 & 32.4 $\pm$ 1.8 & 41.7 $\pm$ 2.4 & 22.4 $\pm$ 1.2 & 23.8 $\pm$ 2.2 & 46.6 $\pm$ 1.0 & 34.0 $\pm$ 1.7 & 35.9 $\pm$ 2.3 \\

    YeastCC & & 66.5 $\pm$ 0.5 $\bullet$ & 64.4 $\pm$ 1.2 & 48.1 $\pm$ 2.1 & 32.4 $\pm$ 1.0 & 45.5 $\pm$ 0.9 & 27.6 $\pm$ 1.1 & 31.3 $\pm$ 2.0 & 62.0 $\pm$ 1.5 & 35.4 $\pm$ 0.8 & 33.3 $\pm$ 2.5 \\

    YeastBP & & 43.8 $\pm$ 0.5 $\bullet$ & 40.9 $\pm$ 2.2 & 32.8 $\pm$ 1.2 & 16.0 $\pm$ 0.6 & 25.5 $\pm$ 0.7 & 16.0 $\pm$ 1.1 & 15.8 $\pm$ 1.0 & 39.9 $\pm$ 1.4 & 17.6 $\pm$ 0.7 & 30.8 $\pm$ 1.3 \\

    \midrule

    \multirow{3}{*}{Scene} & 1 & 86.2 $\pm$ 0.6 $\bullet$ & 84.9 $\pm$ 1.0 & 67.7 $\pm$ 2.9 & 75.1 $\pm$ 1.4 & 76.1 $\pm$ 1.3 & 82.1 $\pm$ 1.3 & 82.2 $\pm$ 1.0 & 77.1 $\pm$ 2.0 & 75.2 $\pm$ 1.4 & 74.7 $\pm$ 2.9 \\ 

    & 2 & 85.8 $\pm$ 1.3 $\bullet$ & 84.0 $\pm$ 1.7 & 67.4 $\pm$ 2.0 & 67.9 $\pm$ 1.6 & 69.4 $\pm$ 1.8 & 79.8 $\pm$ 1.5 & 79.6 $\pm$ 1.2 & 69.6 $\pm$ 2.3 & 68.1 $\pm$ 1.5 & 67.7 $\pm$ 1.5 \\ 

    & 3 & 84.4 $\pm$ 0.9 $\bullet$ & 63.9 $\pm$ 4.1 & 65.9 $\pm$ 2.8 & 60.2 $\pm$ 0.6 & 61.2 $\pm$ 0.6 & 72.3 $\pm$ 1.3 & 73.5 $\pm$ 1.0 & 60.4 $\pm$ 1.1 & 60.4 $\pm$ 0.7 & 60.0 $\pm$ 0.8 \\ 

    \midrule

    \multirow{3}{*}{Birds} & 3 & 61.8 $\pm$ 0.6 $\bullet$ & 55.8 $\pm$ 1.1 & 38.7 $\pm$ 3.8 & 51.3 $\pm$ 5.6 & 54.0 $\pm$ 2.7 & 38.7 $\pm$ 3.9 & 38.3 $\pm$ 1.9 & 46.3 $\pm$ 4.5 & 53.9 $\pm$ 4.9 & 37.8 $\pm$ 6.4 \\ 

    & 7 & 51.9 $\pm$ 0.8 $\bullet$ & 48.4 $\pm$ 1.5 & 35.1 $\pm$ 3.7 & 30.0 $\pm$ 2.3 & 35.1 $\pm$ 3.5 & 28.7 $\pm$ 1.5 & 34.1 $\pm$ 2.9 & 44.9 $\pm$ 3.8 & 29.7 $\pm$ 3.2 & 33.2 $\pm$ 3.3 \\ 

    & 11 & 43.1 $\pm$ 3.2 $\bullet$ & 41.1 $\pm$ 3.7 & 31.4 $\pm$ 3.4 & 27.8 $\pm$ 1.7 & 30.2 $\pm$ 3.7 & 23.2 $\pm$ 4.8 & 27.2 $\pm$ 2.8 & 35.2 $\pm$ 4.0 & 28.6 $\pm$ 2.5 & 28.6 $\pm$ 2.7 \\ 

    \midrule

    \multirow{3}{*}{Medical} & 3 & 90.8 $\pm$ 1.0 $\bullet$ & 87.4 $\pm$ 2.5 & 83.2 $\pm$ 1.7 & 77.7 $\pm$ 1.1 & 87.6 $\pm$ 1.2 & 66.2 $\pm$ 3.3 & 74.5 $\pm$ 6.0 & 61.4 $\pm$ 2.5 & 79.5 $\pm$ 1.2 & 85.2 $\pm$ 1.4 \\ 

    & 7 & 87.1 $\pm$ 0.3 $\bullet$ & 85.8 $\pm$ 1.5 & 72.2 $\pm$ 3.0 & 51.0 $\pm$ 2.5 & 82.8 $\pm$ 1.5 & 64.3 $\pm$ 1.9 & 69.5 $\pm$ 5.1 & 60.1 $\pm$ 2.4 & 65.7 $\pm$ 2.1 & 81.1 $\pm$ 2.2 \\ 

    & 11 & 86.3 $\pm$ 0.7 $\bullet$ & 84.2 $\pm$ 0.8 & 71.5 $\pm$ 2.5 & 38.1 $\pm$ 3.2 & 77.6 $\pm$ 2.0 & 62.3 $\pm$ 4.3 & 70.3 $\pm$ 1.2 & 59.5 $\pm$ 3.5 & 55.3 $\pm$ 4.3 & 77.3 $\pm$ 1.7 \\ 

    \midrule

    \multirow{3}{*}{Enron} & 3 & 70.6 $\pm$ 1.1 $\bullet$ & 64.2 $\pm$ 1.4 & 58.6 $\pm$ 1.3 & 45.1 $\pm$ 1.6 & 60.6 $\pm$ 2.0 & 59.8 $\pm$ 1.5 & 48.8 $\pm$ 2.4 & 67.6 $\pm$ 1.3 & 47.0 $\pm$ 1.9 & 66.5 $\pm$ 1.8 \\ 

    & 7 & 70.5 $\pm$ 0.6 $\bullet$ & 62.0 $\pm$ 1.8 & 58.6 $\pm$ 2.2 & 33.9 $\pm$ 1.0 & 54.6 $\pm$ 2.0 & 58.4 $\pm$ 1.6 & 43.3 $\pm$ 4.4 & 66.4 $\pm$ 1.7 & 39.0 $\pm$ 1.3 & 62.9 $\pm$ 2.2 \\ 

    & 11 & 66.8 $\pm$ 1.0 $\bullet$ & 57.8 $\pm$ 1.9 & 57.9 $\pm$ 1.8 & 27.4 $\pm$ 1.2 & 48.8 $\pm$ 2.2 & 54.8 $\pm$ 2.1 & 41.2 $\pm$ 6.3 & 65.6 $\pm$ 1.8 & 33.3 $\pm$ 1.1 & 59.7 $\pm$ 2.2 \\ 

    \midrule

    \multirow{3}{*}{Chess} & 10 & 47.5 $\pm$ 0.8 $\bullet$ & 44.7 $\pm$ 1.3 & 42.9 $\pm$ 1.2 & 28.9 $\pm$ 1.0 & 34.6 $\pm$ 0.8 & 25.6 $\pm$ 1.4 & 24.6 $\pm$ 1.1 & 42.1 $\pm$ 1.4 & 26.6 $\pm$ 1.0 & 43.7 $\pm$ 0.9 \\ 

    & 20 & 44.7 $\pm$ 0.8 $\bullet$ & 40.0 $\pm$ 1.4 & 40.6 $\pm$ 0.9 & 19.2 $\pm$ 0.7 & 28.3 $\pm$ 0.7 & 25.0 $\pm$ 0.6 & 22.1 $\pm$ 2.1 & 40.0 $\pm$ 1.2 & 20.1 $\pm$ 0.7 & 39.3 $\pm$ 0.6 \\ 

    & 30 & 42.4 $\pm$ 0.4 $\bullet$ & 41.1 $\pm$ 1.4 & 38.5 $\pm$ 1.9 & 14.4 $\pm$ 0.6 & 23.4 $\pm$ 0.9 & 23.9 $\pm$ 1.1 & 16.5 $\pm$ 2.7 & 38.1 $\pm$ 1.4 & 15.5 $\pm$ 0.7 & 35.9 $\pm$ 1.2 \\ 

    \midrule

    \multirow{3}{*}{Philosophy} & 10 & 47.7 $\pm$ 0.9 $\bullet$ & 46.3 $\pm$ 1.3 & 41.9 $\pm$ 1.1 & 31.7 $\pm$ 1.2 & 35.5 $\pm$ 1.0 & 27.3 $\pm$ 0.6 & 26.7 $\pm$ 1.3 & 45.1 $\pm$ 0.8 & 31.5 $\pm$ 1.1 & 44.4 $\pm$ 0.8 \\ 

    & 20 & 44.9 $\pm$ 0.3 $\bullet$ & 42.4 $\pm$ 1.2 & 39.6 $\pm$ 1.0 & 24.4 $\pm$ 1.3 & 29.3 $\pm$ 0.9 & 26.0 $\pm$ 0.6 & 27.5 $\pm$ 1.9 & 41.0 $\pm$ 0.9 & 24.4 $\pm$ 1.3 & 40.4 $\pm$ 0.7 \\ 

    & 30 & 42.5 $\pm$ 0.4 $\bullet$ & 41.2 $\pm$ 1.1 & 37.9 $\pm$ 1.1 & 20.0 $\pm$ 1.5 & 24.7 $\pm$ 1.0 & 23.1 $\pm$ 1.5 & 26.0 $\pm$ 0.9 & 37.3 $\pm$ 0.6 & 20.1 $\pm$ 1.5 & 37.5 $\pm$ 0.7 \\ 

    \bottomrule
\end{tabular}
}
\label{tab:average precision}
\end{table*}

\begin{table*}[!ht]
\renewcommand\arraystretch{0.9} 
\centering
\caption{Experimental results on \textit{ranking loss} (\%, lower is better $\downarrow$). $\bullet$ denotes the best result among all methods.}
\resizebox{\textwidth}{!}{
\begin{tabular}{c c l l l l l l l l l l}
    \toprule
    Data & r & \multicolumn{1}{c}{Schirn} & \multicolumn{1}{c}{NLR} & \multicolumn{1}{c}{FPML} & \multicolumn{1}{c}{PML-LRS} & \multicolumn{1}{c}{PML-NI} & \multicolumn{1}{c}{P-MAP} & \multicolumn{1}{c}{P-VLS} & \multicolumn{1}{c}{PAKS} & \multicolumn{1}{c}{GLC} & \multicolumn{1}{c}{PARD} \\
    
    \midrule
     
    Music\_emotion & & 23.5 $\pm$ 0.6 $\bullet$ & 26.9 $\pm$ 0.8 & 29.2 $\pm$ 0.6 & 24.2 $\pm$ 0.6 & 24.5 $\pm$ 0.7 & 25.9 $\pm$ 1.3 & 26.8 $\pm$ 0.7 & 24.3 $\pm$ 0.7 & 24.2 $\pm$ 0.7 & 24.6 $\pm$ 0.7 \\

    Music\_style & & 13.2 $\pm$ 0.2 $\bullet$ & 13.7 $\pm$ 0.4 & 17.3 $\pm$ 0.4 & 14.6 $\pm$ 0.6 & 13.9 $\pm$ 0.5 & 14.7 $\pm$ 0.5 & 16.7 $\pm$ 0.6 & 14.1 $\pm$ 0.6 & 14.2 $\pm$ 0.6 & 14.5 $\pm$ 0.7 \\

    YeastMF & & 19.7 $\pm$ 0.8 $\bullet$ & 25.8 $\pm$ 1.4 & 32.0 $\pm$ 2.2 & 32.1 $\pm$ 1.9 & 26.7 $\pm$ 2.1 & 37.0 $\pm$ 1.5 & 41.4 $\pm$ 1.3 & 22.9 $\pm$ 0.6 & 31.0 $\pm$ 2.0 & 29.9 $\pm$ 2.0 \\

    YeastCC & & 14.3 $\pm$ 0.4 $\bullet$ & 15.7 $\pm$ 0.8 & 25.4 $\pm$ 1.6 & 31.2 $\pm$ 1.0 & 23.2 $\pm$ 1.2 & 31.6 $\pm$ 2.1 & 35.1 $\pm$ 1.7 & 16.7 $\pm$ 0.7 & 28.8 $\pm$ 1.1 & 33.1 $\pm$ 1.7 \\

    YeastBP & & 18.8 $\pm$ 0.3 $\bullet$ & 20.0 $\pm$ 0.6 & 25.2 $\pm$ 0.8 & 35.4 $\pm$ 0.7 & 28.5 $\pm$ 0.8 & 36.8 $\pm$ 1.2 & 40.3 $\pm$ 0.7 & 21.0 $\pm$ 0.6 & 34.2 $\pm$ 0.8 & 26.2 $\pm$ 0.9 \\

    \midrule

    \multirow{3}{*}{Scene} & 1 & 7.7 $\pm$ 0.5 $\bullet$ & 9.3 $\pm$ 0.7 & 21.5 $\pm$ 2.5 & 16.7 $\pm$ 0.9 & 15.5 $\pm$ 0.8 & 10.4 $\pm$ 0.6 & 12.3 $\pm$ 0.6 & 15.0 $\pm$ 1.4 & 16.6 $\pm$ 0.9 & 16.8 $\pm$ 0.4 \\ 

    & 2 & 8.0 $\pm$ 0.9 $\bullet$ & 9.8 $\pm$ 1.1 & 22.0 $\pm$ 1.6 & 22.1 $\pm$ 1.3 & 20.6 $\pm$ 1.3 & 12.7 $\pm$ 1.4 & 15.0 $\pm$ 0.9 & 21.0 $\pm$ 1.9 & 22.0 $\pm$ 1.2 & 22.1 $\pm$ 1.3 \\ 

    & 3 & 9.0 $\pm$ 0.9 $\bullet$ & 11.1 $\pm$ 0.8 & 24.1 $\pm$ 2.5 & 29.1 $\pm$ 0.4 & 28.1 $\pm$ 0.5 & 18.7 $\pm$ 0.9 & 20.8 $\pm$ 0.6 & 29.3 $\pm$ 0.9 & 29.0 $\pm$ 0.5 & 29.1 $\pm$ 0.6 \\ 

    \midrule

    \multirow{3}{*}{Birds} & 3 & 16.9 $\pm$ 0.8 $\bullet$ & 22.1 $\pm$ 1.4 & 32.6 $\pm$ 3.8 & 24.6 $\pm$ 4.8 & 22.2 $\pm$ 3.6 & 32.3 $\pm$ 2.3 & 36.0 $\pm$ 1.5 & 27.8 $\pm$ 4.1 & 22.8 $\pm$ 4.4 & 34.0 $\pm$ 3.2 \\ 

    & 7 & 25.8 $\pm$ 1.5 $\bullet$ & 28.6 $\pm$ 1.2 & 35.0 $\pm$ 3.3 & 40.4 $\pm$ 2.6 & 34.6 $\pm$ 4.4 & 38.1 $\pm$ 1.8 & 39.7 $\pm$ 3.6 & 30.0 $\pm$ 2.7 & 39.2 $\pm$ 3.2 & 36.9 $\pm$ 4.0 \\ 

    & 11 & 29.1 $\pm$ 3.3 $\bullet$ & 34.4 $\pm$ 2.9 & 39.5 $\pm$ 2.5 & 43.0 $\pm$ 1.2 & 39.2 $\pm$ 2.2 & 47.3 $\pm$ 3.1 & 48.0 $\pm$ 3.5 & 35.3 $\pm$ 1.7 & 42.0 $\pm$ 1.6 & 40.2 $\pm$ 1.6 \\ 

    \midrule

    \multirow{3}{*}{Medical} & 3 & 2.2 $\pm$ 0.6 $\bullet$ & 3.4 $\pm$ 1.2 & 4.6 $\pm$ 1.3 & 5.1 $\pm$ 0.5 & 2.6 $\pm$ 0.9 & 9.1 $\pm$ 1.3 & 10.3 $\pm$ 2.6 & 9.4 $\pm$ 1.0 & 4.6 $\pm$ 0.7 & 3.1 $\pm$ 0.9 \\ 

    & 7 & 3.5 $\pm$ 0.5 $\bullet$ & 4.6 $\pm$ 1.1 & 9.5 $\pm$ 1.2 & 13.9 $\pm$ 1.3 & 4.3 $\pm$ 1.0 & 10.2 $\pm$ 0.6 & 11.0 $\pm$ 1.1 & 11.3 $\pm$ 0.5 & 8.6 $\pm$ 1.1 & 4.7 $\pm$ 1.2 \\ 

    & 11 & 3.9 $\pm$ 0.6 $\bullet$ & 5.6 $\pm$ 1.1 & 9.6 $\pm$ 1.6 & 19.5 $\pm$ 1.8 & 5.4 $\pm$ 0.8 & 11.3 $\pm$ 1.3 & 10.7 $\pm$ 1.4 & 11.9 $\pm$ 1.5 & 11.4 $\pm$ 1.8 & 5.8 $\pm$ 0.7 \\ 

    \midrule

    \multirow{3}{*}{Enron} & 3 & 8.8 $\pm$ 0.5 $\bullet$ & 12.9 $\pm$ 0.5 & 13.9 $\pm$ 2.3 & 22.5 $\pm$ 1.6 & 14.6 $\pm$ 1.2 & 11.7 $\pm$ 0.5 & 18.5 $\pm$ 0.6 & 9.9 $\pm$ 0.7 & 22.4 $\pm$ 1.7 & 11.0 $\pm$ 1.0 \\ 

    & 7 & 9.0 $\pm$ 0.3 $\bullet$ & 14.1 $\pm$ 1.1 & 14.9 $\pm$ 2.0 & 27.7 $\pm$ 1.1 & 17.1 $\pm$ 1.4 & 12.4 $\pm$ 0.4 & 19.1 $\pm$ 0.7 & 11.1 $\pm$ 0.8 & 25.4 $\pm$ 1.4 & 13.2 $\pm$ 1.3 \\ 

    & 11 & 12.0 $\pm$ 0.4 & 16.4 $\pm$ 1.6 & 15.6 $\pm$ 0.9 & 31.8 $\pm$ 0.8 & 19.6 $\pm$ 1.5 & 14.0 $\pm$ 0.3 & 18.6 $\pm$ 1.4 & \textbf{11.7} $\pm$ 0.8 $\bullet$ & 28.1 $\pm$ 1.1 & 14.8 $\pm$ 1.4 \\ 

    \midrule

    \multirow{3}{*}{Chess} & 10 & 12.6 $\pm$ 0.7 $\bullet$ & 14.4 $\pm$ 0.4 & 15.0 $\pm$ 1.0 & 24.3 $\pm$ 1.1 & 20.0 $\pm$ 1.1 & 19.7 $\pm$ 1.1 & 24.3 $\pm$ 1.7 & 14.9 $\pm$ 0.8 & 25.6 $\pm$ 1.0 & 13.6 $\pm$ 0.9 \\ 

    & 20 & 14.7 $\pm$ 0.1 $\bullet$ & 16.2 $\pm$ 1.1 & 16.8 $\pm$ 0.7 & 29.0 $\pm$ 1.0 & 23.5 $\pm$ 1.0 & 21.6 $\pm$ 1.5 & 24.0 $\pm$ 1.9 & 17.1 $\pm$ 1.1 & 28.9 $\pm$ 1.1 & 17.0 $\pm$ 0.9 \\ 

    & 30 & 15.8 $\pm$ 0.3 $\bullet$ & 18.1 $\pm$ 1.5 & 18.4 $\pm$ 0.9 & 31.7 $\pm$ 0.7 & 25.9 $\pm$ 0.8 & 23.8 $\pm$ 1.2 & 26.3 $\pm$ 1.5 & 19.5 $\pm$ 0.9 & 31.5 $\pm$ 0.8 & 18.6 $\pm$ 0.8 \\ 

    \midrule

    \multirow{3}{*}{Philosophy} & 10 & 13.2 $\pm$ 0.4 $\bullet$ & 14.3 $\pm$ 0.7 & 15.9 $\pm$ 0.8 & 24.0 $\pm$ 0.7 & 21.0 $\pm$ 0.9 & 19.9 $\pm$ 0.5 & 19.9 $\pm$ 0.9 & 16.0 $\pm$ 1.1 & 24.7 $\pm$ 0.7 & 13.4 $\pm$ 0.8 \\ 

    & 20 & 15.9 $\pm$ 0.3 $\bullet$ & 17.7 $\pm$ 0.9 & 18.2 $\pm$ 0.8 & 27.1 $\pm$ 0.4 & 24.0 $\pm$ 0.6 & 20.9 $\pm$ 0.8 & 21.4 $\pm$ 0.9 & 18.8 $\pm$ 1.0 & 27.5 $\pm$ 0.5 & 16.5 $\pm$ 1.0 \\ 

    & 30 & 17.8 $\pm$ 0.3 $\bullet$ & 19.3 $\pm$ 1.1 & 20.2 $\pm$ 1.0 & 29.2 $\pm$ 0.5 & 25.9 $\pm$ 0.5 & 24.1 $\pm$ 0.8 & 23.2 $\pm$ 1.1 & 20.7 $\pm$ 1.0 & 29.5 $\pm$ 0.5 & 18.4 $\pm$ 0.8 \\ 
    
    \bottomrule
\end{tabular}
}
\label{tab:ranking loss}
\end{table*}

\begin{table*}[!ht]
\renewcommand\arraystretch{1} 
    \centering
    \caption{Win/tie/loss counts on the multi-label metrics of Schirn against each comparing method on real-world data sets and synthetic data sets according to the pairwise $t$-test at 0.05 significance level.}
    \resizebox{\textwidth}{!}{
    \begin{tabular}{ c c c c c c c c c c c }
        \toprule
        
         & \multicolumn{1}{c}{NLR} & \multicolumn{1}{c}{FPML} & \multicolumn{1}{c}{PML-LRS} & \multicolumn{1}{c}{PML-NI} & \multicolumn{1}{c}{P-MAP} & \multicolumn{1}{c}{P-VLS} & \multicolumn{1}{c}{PAKS} & \multicolumn{1}{c}{GLC} & \multicolumn{1}{c}{PARD} & Total \\ 
        
        \midrule
        Average Precision & 23/0/0 & 23/0/0 & 23/0/0 & 23/0/0 & 23/0/0 & 23/0/0 & 23/0/0 & 23/0/0 & 23/0/0 & 207/0/0 \\
        Ranking Loss & 23/0/0 & 23/0/0 & 22/1/0 & 22/1/0 & 23/0/0 & 21/1/1 & 22/1/0 & 22/1/0 & 20/3/0 & 198/8/1 \\
        Coverage & 17/2/4 & 20/0/3 & 19/1/3 & 17/3/3 & 20/0/3 & 19/1/3 & 19/1/3 & 19/1/3 & 20/0/3 & 170/9/28 \\ 
         Hamming Loss & 17/6/0 & 19/4/0 & 23/0/0 & 22/1/0 & 23/0/0 & 19/4/0 & 20/0/3 & 22/1/0 & 17/5/1 & 182/21/4 \\
         One-error & 23/0/0 & 23/0/0 & 22/1/0 & 23/0/0 & 23/0/0 & 22/0/1 & 23/0/0 & 23/0/0 & 23/0/0 & 205/1/1 \\
         \midrule
         Total & 103/8/4 & 108/4/3 & 109/3/3 & 107/5/3 & 112/0/3 & 104/6/5 & 107/2/6 & 109/3/3 & 103/8/4 & 962/39/34 \\
        
        \bottomrule
    \end{tabular}
    }       
    \label{tab:ttest}
\end{table*}

To further quantify its performance, we conducted pairwise t-tests at a significance level of 0.05, reporting the win/tie/loss counts of Schirn against each comparing method across the five metrics in Table. \ref{tab:ttest}. 
Schirn achieves win rates of 100\%, 95.7\%, 82.1\%, 87.9\%, and 99.0\% on average precision, ranking loss, coverage, hamming loss, and one-error, respectively. 
These results confirm that Schirn is significantly better than competing methods in the vast majority of cases. 
Overall, Schirn surpasses other methods in 92.9\% of the evaluated cases, demonstrating its robustness across diverse datasets, varying label noise levels, and different evaluation criteria.

\subsection{Further Analysis}

\subsubsection{Ablation Study}

To assess the contributions of key components in Schirn, we conduct ablation studies on six synthetic datasets (Tables~\ref{tab:ablation} and \ref{tab:ablation1}–\ref{tab:ablation2}, Appendix).

\begin{table*}[!ht]
\renewcommand\arraystretch{1.2} 
\centering
\caption{Ablation study of Schirn in terms of \textit{average precision} and \textit{ranking loss} on synthetic datasets (in \%). The settings for each synthetic dataset are $r=1$ for Scene, $r=3$ for Birds, Medical, and Enron, and $r=10$ for Chess and Philosophy.}
\resizebox{\textwidth}{!}{
\begin{tabular}{ c c c | c c c c c c | c c c c c c }
    \toprule
    
    High & \multirow{2}{*}{Sparsity} & Low & \multicolumn{6}{c|}{\textbf{average precision $\uparrow$}} & \multicolumn{6}{c}{\textbf{ranking loss $\downarrow$}} \\ 
    Rank & & Rank & Scene & Birds & Medical & Enron & Chess & Philosophy & Scene & Birds & Medical & Enron & Chess & Philosophy \\

    \midrule

    \ding{53} & \checkmark & \ding{53} & 83.1 $\pm$ 1.0 & 53.3 $\pm$ 4.1 & 88.4 $\pm$ 1.6 & 68.6 $\pm$ 1.6 & 43.3 $\pm$ 1.4 & 44.9 $\pm$ 0.8 & 9.8 $\pm$ 0.8 & 22.6 $\pm$ 2.9 & 3.3 $\pm$ 1.0 & 10.3 $\pm$ 0.4 & 13.7 $\pm$ 0.8 & 14.6 $\pm$ 0.4 \\

    \checkmark & \ding{53} & \ding{53} & 58.2 $\pm$ 1.7 & 44.2 $\pm$ 2.0 & 84.5 $\pm$ 1.7 & 47.0 $\pm$ 1.6 & 3.2 $\pm$ 1.0 & 12.4 $\pm$ 1.8 & 33.0 $\pm$ 2.6 & 34.4 $\pm$ 2.3 & 6.2 $\pm$ 0.8 & 29.4 $\pm$ 2.0 & 47.0 $\pm$ 5.5 & 45.2 $\pm$ 2.9 \\

    \ding{53} & \checkmark &  \checkmark & 83.7 $\pm$ 1.2 & 53.7 $\pm$ 5.4 & 88.4 $\pm$ 1.8 & 68.0 $\pm$ 0.7 & 43.1 $\pm$ 1.3 & 45.5 $\pm$ 0.6 & 9.2 $\pm$ 0.9 & 23.5 $\pm$ 3.8 & 3.4 $\pm$ 0.9 & 10.5 $\pm$ 0.6 & 14.0 $\pm$ 1.2 & 14.7 $\pm$ 0.2 \\

    \hline

    \checkmark & \checkmark & \ding{53} & 86.2 $\pm$ 0.6 & 61.8 $\pm$ 0.6 & 90.8 $\pm$ 1.0 & 70.6 $\pm$ 1.1 & 47.5 $\pm$ 0.8 & 47.7 $\pm$ 0.9 & 7.7 $\pm$ 0.5 & 16.9 $\pm$ 0.8 & 2.2 $\pm$ 0.6 & 8.8 $\pm$ 0.5 & 12.6 $\pm$ 0.7 & 13.2 $\pm$ 0.4 \\

    \bottomrule
\end{tabular}
}       
\label{tab:ablation}
\end{table*}

\textbf{High-rank Property}: 
By comparing the rows with and without the high-rank property, we can observe that enforcing a high-rank constraint consistently improves the performance across all datasets. 
For instance, on Birds, the average precision improves significantly from 0.533 to 0.618.
These results validate that preserving the high-rank property of the predicted label matrix is crucial for maintaining the richness and complexity of label correlations.

\textbf{Sparsity Constraint}:
The sparsity constraint on the noise label matrix also plays a vital role in improving model performance. 
When the sparsity constraint is removed, the performance degrade significantly. 
For example, on dataset Enron, removing the sparsity constraint reduces the average precision from 0.706 to 0.470 and increases the ranking loss from 0.088 to 0.294. 
This highlights that the sparsity constraint effectively identifies and suppresses noisy labels, thereby reducing generalization error.

\textbf{High-rank vs. Low-rank}:
To further examine the assumptions of the high-rank and low-rank property, we also evaluate the performance when a low-rank constraint is applied. 
While the low-rank constraint shows some effects, it is consistently outperformed by the high-rank constraint. 
For instance, on the Chess dataset, the ranking loss is lower under the high-rank constraint (0.126 vs. 0.140). 
These findings reaffirm that the true label matrix is generally full-rank or near full-rank, and enforcing a high-rank constraint better aligns with the inherent structure of the data.

\subsubsection{Does Schirn Preserve the High-Rank Property of the Label Matrix?}

\begin{wraptable}{r}{7.5cm}
\centering
\caption{The rank of the label matrix across real-world partial multi-label datasets. Here, $r(\mathbf{P})$, $\beta=0$, $r(\mathbf{Y})$ and $r(\mathbf{Y}_g)$ represent the rank of prediction label matrix, the rank of that when $\beta=0$, the rank of observed label matrix and true label matrix, respectively.}
\begin{tabular}{c c c c c}
    \toprule
        & $r(\mathbf{P})$ & $\beta=0$ & $r(\mathbf{Y}_g)$ & $r(\mathbf{Y})$ \\ 
        \midrule
         Music\_emotion & 11 & 8 & 11 & 11 \\
         Music\_style & 10 & 10 & 10 & 10 \\
         YeastMF & 38 & 24 & 36 & 39 \\
         YeastCC & 45 & 23 & 43 & 50 \\
         YeastBP & 210 & 115 & 182 & 217 \\
        \bottomrule
    \end{tabular}
\label{tab:rank}
\end{wraptable} 

To evaluate whether Schirn preserves the high-rank property of the predicted label matrix, we compare the rank of the prediction $\mathbf{P}$, denoted $r(\mathbf{P})$, with that of the observed label matrix $r(\mathbf{Y})$ and the ground-truth matrix $r(\mathbf{Y}_g)$ across real-world PML datasets. 
We also assess the effect of removing the high-rank term (i.e., setting $\beta = 0$). 
As shown in Table. \ref{tab:rank}, Schirn produces a prediction matrix whose rank closely aligns with the ground-truth, while substantially filtering out the noise inherent in $\mathbf{Y}$. 
In contrast, omitting the high-rank term leads to a noticeable drop in matrix rank. 
These results demonstrate Schirn’s ability to preserve the structural richness of label correlations while maintaining robustness against noisy labels.


\subsubsection{Parameter Sensitivity and Convergence}

\begin{figure*}[ht!]
    \centering
    \includegraphics[width=1\linewidth]{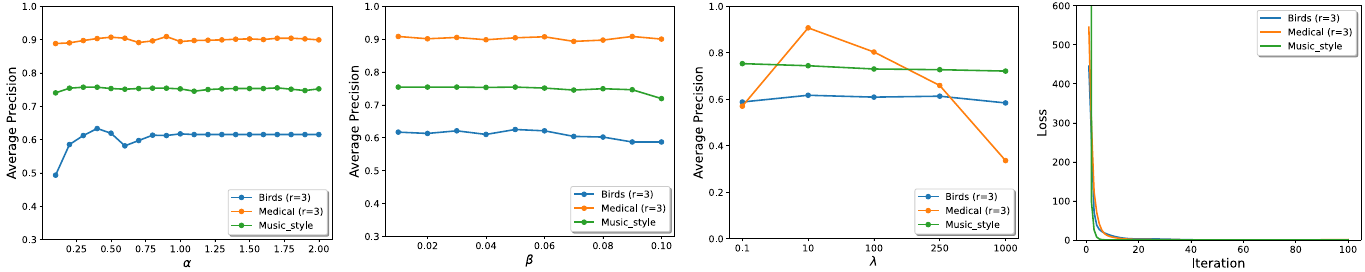}
    \caption{Sensitivity analysis of Schirn.}
    \label{fig:sensitivity}
\end{figure*}

We evaluate the parameter sensitivity and convergence of Schirn, as illustrated in the Fig. \ref{fig:sensitivity}. 
It is evident that within a reasonable range, Schirn demonstrates robustness to parameter variations. 
Additionally, Schirn exhibits rapid convergence within a few epochs, showcasing its efficiency. 
These properties highlight the advantages of Schirn, making it well-suited for real-world applications.

%% file: sec/6.conclusion.tex
\section{Conclusion}
In this paper, we propose \textbf{Schirn}, a novel approach to Partial Multi-Label Learning (PML) that overcomes the limitations of existing methods, which rely on conflicting assumptions of sparsity in noisy label matrices and low-rankness in true label matrices.
We theoretically show that the rank of a full-rank matrix remains high under sparse perturbations, forming the foundation of our method.
Schirn integrates these insights into a unified framework, enforcing sparsity on the noise label matrix while preserving the high-rank property of the predicted label matrix.
This dual constraint enables effective disambiguation of noisy labels while maintaining the richness and complexity of real-world label structures.
Extensive experiments on real-world and synthetic datasets demonstrate that Schirn achieves superior performance across multiple metrics compared to state-of-the-art methods.
To our knowledge, Schirn is the first work to explore the relationship between sparsity and high-rank properties in PML, advancing label structure modeling and providing a framework for future research in multi-label and weakly supervised learning paradigms.

%% file: sec/appendix.tex
\newpage
\appendix

\section{Proof of Theorem \ref{theorem}} \label{sec appendix proof}
We here prove the theorem \ref{theorem} as follows.

By definition, the noise matrix $\mathbf{N}$ satisfies $\|\mathbf{N}\|_0 \leq \epsilon$, which implies that it contains at most $\epsilon$ non-zero entries.
Sparse binary matrices can be approximated as low-rank matrices since their rank is constrained by the number of non-zero entries:
\begin{equation}
    \text{rank}(\mathbf{N}) \leq \min(\epsilon, n, l).
\end{equation}
The rank of the difference of two matrices $\mathbf{Y}$ and $\mathbf{N}$ is bounded by the following rank inequality:
\begin{equation}
     \text{rank}(\mathbf{Y}_g) = \text{rank}(\mathbf{Y} - \mathbf{N}) \geq \text{rank}(\mathbf{Y}) - \text{rank}(\mathbf{N}).
\end{equation}
Since $\mathbf{Y}$ is full rank with $\text{rank}(\mathbf{Y}) = \min(n, l)$ and $\text{rank}(\mathbf{N}) \leq \epsilon$, it follows that:
\begin{equation}
    \text{rank}(\mathbf{Y}_g) \geq \min(n, l) - \epsilon.
\end{equation}

One of our key assumptions is that $\epsilon$ is a small value that is smaller than $n$ and $l$.
Indeed, many real-world datasets have the property of $n>\epsilon$, such as the original YeastMF, YeastBP, and YeastCC. 
Besides, regarding synthetic datasets, the approach described in \cite{sun2019partial} can also yield extremely sparse noise labels—that is, $\epsilon$ is much smaller than $n$.
Moreover, consider more realistic scenarios. 
For high quality real-world multi-label learning dataset like RCV1-V2, and large scale dataset Amazon-titles with more than 1.3M labels, due to industrial-grade quality control, the noise inevitably introduced during manual labeling but tends to be very sparse, which naturally aligns with our assumption $\epsilon < l$.

Therefore, based on the characteristics of widely-used MLL datasets, we find that many cases satisfy $\epsilon <n, l$. Our use of this assumption is primarily motivated by the desire to establish a sound theoretical analysis within a constrained and analyzable framework.




\section{Details of Datasets}
We here provide the details of datasets used in our experiments in Table \ref{tab:dataset_details}.

\begin{table*}[ht!]
    \centering
    \renewcommand{\arraystretch}{1}
    \caption{Characteristics of real-world datasets. Here, $n$ represents the number of samples, $d$ represents the feature dimensions, and $l$ represents the number of labels. Avg. CLs and Avg. GLs represent the average number of candidate labels per sample and that of ground-truth labels per sample, respectively.}
    \begin{tabular}{l c c c c c c}
        \toprule
       Type & Dataset & $n$ & $d$ & $l$ & Avg. CLs & Avg. GLs \\ 
        \midrule
        \multirow{5}{*}{\textbf{Real-world Data Set}} 
            & Music\_emotion & 6833 & 98 & 11 & 5.29 & 2.42 \\
            & Music\_style & 6839 & 98 & 10 & 6.04 & 1.44 \\
            & YeastMF & 1408 & 6139 & 39 & 6.54 & 3.54 \\
            & YeastCC & 1771 & 6139 & 50 & 9.30 & 4.30 \\
            & YeastBP & 3794 & 6139 & 217 & 18.84 & 8.84 \\
        \midrule
        \multirow{6}{*}{\textbf{Synthetic Data Set}} 
            & Scene & 2407 & 294 & 6 & - & 1.07 \\
            & Birds & 645 & 260 & 19 & - & 1.01 \\
            & Medical & 978 & 1449 & 45 & - & 1.25 \\
            & Enron & 1702 & 1001 & 53 & - & 3.38 \\
            & Chess & 1675 & 585 & 227 & - & 2.41 \\
            & Philosophy & 3971 & 842 & 233 & - & 2.27 \\
        \bottomrule
    \end{tabular}
    \label{tab:dataset_details}
\end{table*}

\section{Co-occurrence Induced Low-rank Label Matrix}
Co-occurrence is a commonly observed phenomenon in multi-label learning (MLL) settings. 
Specifically, when a label A always appears together with another label B—i.e., if a sample has label A, it must also have label B—this pattern is referred to as label co-occurrence. 
Such co-occurrence structures naturally induce a low-rank property in the label matrix, as redundant correlations between labels reduce its effective dimensionality.

To examine the robustness of our method under such conditions, we constructed three variants of the Yeast dataset where the ground-truth label matrix is explicitly made low-rank by introducing artificial co-occurrence relationships among labels. The experimental setup follows the protocol introduced in \cite{si2023multi}. 
We then evaluated three approaches: our proposed method, our method augmented with a low-rank constraint, and NLR \cite{yang2024noisy}.

As reported in Table~\ref{tab:yeast_rank_ablation}, our method consistently and significantly outperforms the other two baselines across all metrics, demonstrating its superior performance and resilience even when the label space exhibits strong low-rank characteristics.

\begin{table*}[!ht]
\renewcommand\arraystretch{1.1}
\centering
\caption{Comparison of high-rank, low-rank, and NLR settings on Yeast datasets. We report the rank of the label matrix (original and after transformation), average precision (↑), one-error (↓), ranking loss (↓), coverage (↓), and hamming loss (↓), all in percentage format.}
\resizebox{\textwidth}{!}{
\begin{tabular}{l l c c c c c c c}
\toprule
Dataset & Condition & rank original & rank new & Avg. Precision $\uparrow$ & One-error $\downarrow$ & Ranking Loss $\downarrow$ & Coverage $\downarrow$ & Hamming Loss $\downarrow$ \\
\midrule

\multirow{3}{*}{Yeast-MF} 
& High-rank & 36 & 29 & 51.2 $\pm$ 1.7 & 48.5 $\pm$ 2.8 & 24.3 $\pm$ 1.4 & 43.4 $\pm$ 2.1 & 11.7 $\pm$ 0.5 \\
& Low-rank  & -- & -- & 48.8 $\pm$ 1.3 & 50.1 $\pm$ 2.2 & 26.6 $\pm$ 1.6 & 46.9 $\pm$ 1.9 & 11.9 $\pm$ 0.6 \\
& NLR       & -- & -- & 25.5 $\pm$ 1.1 & 66.0 $\pm$ 3.5 & 46.1 $\pm$ 1.7 & 72.7 $\pm$ 2.2 & 13.1 $\pm$ 0.6 \\
\midrule

\multirow{3}{*}{Yeast-CC} 
& High-rank & 45 & 37 & 67.3 $\pm$ 2.1 & 30.6 $\pm$ 2.7 & 15.2 $\pm$ 1.7 & 35.6 $\pm$ 2.0 & 11.4 $\pm$ 0.1 \\
& Low-rank  & -- & -- & 65.1 $\pm$ 2.0 & 32.4 $\pm$ 3.0 & 16.4 $\pm$ 1.7 & 37.9 $\pm$ 1.8 & 10.9 $\pm$ 0.0 \\
& NLR       & -- & -- & 24.3 $\pm$ 2.2 & 72.1 $\pm$ 2.4 & 44.1 $\pm$ 1.8 & 76.9 $\pm$ 1.8 & 13.8 $\pm$ 0.3 \\
\midrule

\multirow{3}{*}{Yeast-BP} 
& High-rank & 200 & 158 & 44.2 $\pm$ 1.0 & 47.5 $\pm$ 1.6 & 20.1 $\pm$ 0.2 & 48.7 $\pm$ 0.7 & 5.5 $\pm$ 0.1 \\
& Low-rank  & -- & -- & 41.6 $\pm$ 1.0 & 49.2 $\pm$ 2.0 & 22.2 $\pm$ 0.3 & 53.4 $\pm$ 1.2 & 5.4 $\pm$ 0.1 \\
& NLR       & -- & -- & 8.8 $\pm$ 1.0 & 88.5 $\pm$ 2.1 & 49.3 $\pm$ 4.2 & 88.8 $\pm$ 3.7 & 6.0 $\pm$ 0.1 \\

\bottomrule
\end{tabular}
}
\label{tab:yeast_rank_ablation}
\end{table*}

\section{Limitations}
One limitation of this work is that our method has only been evaluated in the context of stand-alone frameworks, without integration into broader deep learning architectures.
While the proposed Schirn formulation is designed to be modular and broadly applicable, its effectiveness in more complex or end-to-end neural architectures remains to be explored. 
In future work, we plan to extend our investigation by applying Schirn to general-purpose deep learning models to further validate its adaptability and more practical utility.

\begin{algorithm}[ht]
   \caption{The pseudo code of Schirn}
   \label{alg:schirn}
\begin{algorithmic}[1]
   \STATE {\bfseries Input:} partial multi-label dataset $\mathcal{D}$, hyper-parameters $\alpha$, $\beta$ and $\lambda$;
   \STATE {\bfseries Output:} The weight matrix $\mathbf{W}$;
   \STATE Initialize $\mathbf{W} = \mathbf{0}_{d\times l}$, $\mathbf{N} = \mathbf{0}_{n\times l}$, $ \mathbf{C} = \mathbf{\Lambda} = \mathbf{1}_{n\times l}$, $\mu=10^{-4}$, $\mu_{\max} = 10$, $\rho=1.1$, and the maximum number of iteration $iter=100$;
   \FOR{$i=0$ to iter}
    \STATE Update $\mathbf{W}$ according to Eq. (\ref{eq W solution});
    \STATE Update $\mathbf{N}$ according to Eq. (\ref{eq N solution});
    \STATE Update $\mathbf{C}$ according to Eq. (\ref{eq C solution});
    \STATE Update $\mathbf{\Lambda}$ according to Eq. (\ref{eq Lambda solution});
    \STATE Update $\mu$ according to Eq. (\ref{eq mu solution});
   \ENDFOR
   \STATE {\bfseries Return Result}.
\end{algorithmic}
\end{algorithm}


\begin{table*}[!ht]
\renewcommand\arraystretch{1} 
\centering
\caption{Experimental results on \textit{coverage} (\%, lower is better $\downarrow$). $\bullet$ denotes the best result among all methods.}
\resizebox{\textwidth}{!}{
\begin{tabular}{c c l l l l l l l l l l}
    \toprule
    Data & r & \multicolumn{1}{c}{Schirn} & \multicolumn{1}{c}{NLR} & \multicolumn{1}{c}{FPML} & \multicolumn{1}{c}{PML-LRS} & \multicolumn{1}{c}{PML-NI} & \multicolumn{1}{c}{P-MAP} & \multicolumn{1}{c}{P-VLS} & \multicolumn{1}{c}{PAKS} & \multicolumn{1}{c}{GLC} & \multicolumn{1}{c}{PARD} \\ 
    
    \midrule
    
    Music\_emotion & & 40.2 $\pm$ 0.7 $\bullet$ & 42.9 $\pm$ 0.7 & 44.9 $\pm$ 0.5 & 40.7 $\pm$ 0.4 & 40.8 $\pm$ 0.3 & 42.0 $\pm$ 1.1 & 41.0 $\pm$ 0.7 & 40.7 $\pm$ 0.4 & 40.6 $\pm$ 0.4 & 41.4 $\pm$ 0.6 \\
    Music\_style & & 19.3 $\pm$ 0.4 $\bullet$ & 19.7 $\pm$ 0.7 & 23.3 $\pm$ 0.5 & 20.7 $\pm$ 0.8 & 19.8 $\pm$ 0.8 & 20.6 $\pm$ 0.7 & 20.7 $\pm$ 0.8 & 19.9 $\pm$ 0.8 & 20.3 $\pm$ 0.8 & 20.6 $\pm$ 0.8 \\
    YeastMF & & 35.3 $\pm$ 0.8 $\bullet$ & 38.2 $\pm$ 0.9 & 49.1 $\pm$ 1.7 & 52.2 $\pm$ 1.7 & 45.0 $\pm$ 2.1 & 54.4 $\pm$ 2.2 & 59.7 $\pm$ 1.9 & 40.2 $\pm$ 0.7 & 52.6 $\pm$ 1.9 & 38.9 $\pm$ 3.1 \\
    YeastCC & & 30.6 $\pm$ 0.6 $\bullet$ & 31.9 $\pm$ 0.9 & 44.0 $\pm$ 2.1 & 54.5 $\pm$ 1.0 & 45.2 $\pm$ 1.6 & 53.4 $\pm$ 3.1 & 61.6 $\pm$ 2.2 & 34.9 $\pm$ 0.8 & 53.0 $\pm$ 1.3 & 47.8 $\pm$ 2.0 \\
    YeastBP & & 42.9 $\pm$ 1.0 $\bullet$ & 44.8 $\pm$ 1.1 & 50.4 $\pm$ 1.5 & 68.2 $\pm$ 1.0 & 60.4 $\pm$ 1.3 & 67.1 $\pm$ 1.0 & 74.3 $\pm$ 1.3 & 48.5 $\pm$ 1.1 & 68.1 $\pm$ 1.1 & 51.5 $\pm$ 1.0 \\

    \midrule

    \multirow{3}{*}{Scene} & 1 & 7.7 $\pm$ 0.3 $\bullet$ & 9.3 $\pm$ 0.7 & 19.5 $\pm$ 2.1 & 15.6 $\pm$ 0.8 & 14.5 $\pm$ 0.8 & 10.2 $\pm$ 0.6 & 10.2 $\pm$ 0.6 & 14.1 $\pm$ 1.3 & 15.5 $\pm$ 0.8 & 15.5 $\pm$ 1.9 \\
    & 2 & 8.1 $\pm$ 0.6 $\bullet$ & 9.7 $\pm$ 1.0 & 19.9 $\pm$ 1.3 & 20.1 $\pm$ 1.1 & 18.8 $\pm$ 1.2 & 12.0 $\pm$ 1.2 & 11.9 $\pm$ 0.7 & 19.1 $\pm$ 1.7 & 20.0 $\pm$ 1.0 & 20.1 $\pm$ 1.3 \\
    & 3 & 9.0 $\pm$ 0.8 $\bullet$ & 10.8 $\pm$ 0.9 & 21.7 $\pm$ 2.2 & 25.9 $\pm$ 0.5 & 25.0 $\pm$ 0.6 & 17.1 $\pm$ 0.5 & 16.0 $\pm$ 0.3 & 26.0 $\pm$ 0.9 & 25.8 $\pm$ 0.5 & 25.9 $\pm$ 0.5 \\

    \midrule

    \multirow{3}{*}{Birds} & 3 & 26.1 $\pm$ 0.5 & 15.6 $\pm$ 3.0 & 21.6 $\pm$ 3.8 & 16.8 $\pm$ 4.1 & 15.6 $\pm$ 3.7 $\bullet$ & 20.7 $\pm$ 2.4 & 21.7 $\pm$ 1.6 & 18.9 $\pm$ 3.6 & 15.7 $\pm$ 4.1 & 22.3 $\pm$ 1.9 \\
    & 7 & 38.8 $\pm$ 2.7 & 19.4 $\pm$ 1.4 & 22.3 $\pm$ 3.2 & 25.3 $\pm$ 3.1 & 22.2 $\pm$ 4.0 & 23.7 $\pm$ 2.7 & 23.9 $\pm$ 3.7 & 20.2 $\pm$ 2.9 $\bullet$ & 24.6 $\pm$ 3.6 & 23.2 $\pm$ 3.8 \\
    & 11 & 40.2 $\pm$ 4.0 & 21.8 $\pm$ 3.2 & 24.8 $\pm$ 2.1 & 26.1 $\pm$ 1.5 & 24.5 $\pm$ 2.0 & 28.3 $\pm$ 1.8 & 28.4 $\pm$ 3.3 & 23.1 $\pm$ 1.8 $\bullet$ & 25.7 $\pm$ 1.2 & 24.5 $\pm$ 2.1 \\

    \midrule

    \multirow{3}{*}{Medical} & 3 & 3.5 $\pm$ 0.8 $\bullet$ & 5.0 $\pm$ 1.3 & 6.6 $\pm$ 1.6 & 6.6 $\pm$ 0.6 & 4.0 $\pm$ 1.2 & 11.7 $\pm$ 1.8 & 11.3 $\pm$ 2.4 & 11.5 $\pm$ 1.4 & 6.0 $\pm$ 0.8 & 4.6 $\pm$ 1.2 \\
    & 7 & 5.1 $\pm$ 0.6 $\bullet$ & 6.6 $\pm$ 1.3 & 11.8 $\pm$ 1.5 & 16.1 $\pm$ 1.2 & 6.2 $\pm$ 1.2 & 13.2 $\pm$ 1.1 & 13.4 $\pm$ 1.4 & 13.4 $\pm$ 0.7 & 10.6 $\pm$ 0.9 & 6.7 $\pm$ 1.4 \\
    & 11 & 5.9 $\pm$ 1.0 $\bullet$ & 7.8 $\pm$ 1.5 & 12.0 $\pm$ 2.0 & 22.1 $\pm$ 1.7 & 7.3 $\pm$ 0.7 & 14.2 $\pm$ 1.5 & 13.5 $\pm$ 2.0 & 14.2 $\pm$ 1.8 & 13.6 $\pm$ 1.7 & 7.8 $\pm$ 0.8 \\

    \midrule

    \multirow{3}{*}{Enron} & 3 & 24.5 $\pm$ 0.6 $\bullet$ & 26.6 $\pm$ 2.1 & 31.2 $\pm$ 1.7 & 46.3 $\pm$ 2.2 & 36.7 $\pm$ 2.8 & 30.6 $\pm$ 1.1 & 38.7 $\pm$ 1.7 & 26.7 $\pm$ 1.8 & 47.3 $\pm$ 2.3 & 29.8 $\pm$ 2.7 \\
    & 7 & 25.7 $\pm$ 0.5 $\bullet$ & 28.1 $\pm$ 2.2 & 33.0 $\pm$ 2.7 & 50.7 $\pm$ 1.6 & 39.9 $\pm$ 2.6 & 32.6 $\pm$ 1.6 & 40.2 $\pm$ 2.8 & 29.2 $\pm$ 1.9 & 49.8 $\pm$ 1.8 & 34.2 $\pm$ 2.7 \\
    & 11 & 32.3 $\pm$ 1.0 & 29.3 $\pm$ 1.2 $\bullet$ & 35.4 $\pm$ 1.4 & 55.0 $\pm$ 1.3 & 42.7 $\pm$ 2.5 & 34.9 $\pm$ 1.2 & 40.4 $\pm$ 1.5 & 30.6 $\pm$ 1.9 & 52.3 $\pm$ 1.5 & 36.7 $\pm$ 2.4 \\

    \midrule

    \multirow{3}{*}{Chess} & 10 & 24.3 $\pm$ 0.9 $\bullet$ & 26.7 $\pm$ 1.2 & 29.9 $\pm$ 1.6 & 41.6 $\pm$ 1.1 & 36.2 $\pm$ 1.2 & 34.7 $\pm$ 2.1 & 40.4 $\pm$ 3.0 & 29.2 $\pm$ 0.6 & 43.2 $\pm$ 1.2 & 26.9 $\pm$ 0.8 \\
    & 20 & 27.6 $\pm$ 0.6 $\bullet$ & 29.2 $\pm$ 1.4 & 32.0 $\pm$ 1.6 & 46.5 $\pm$ 1.2 & 40.3 $\pm$ 0.9 & 38.3 $\pm$ 1.5 & 41.2 $\pm$ 2.7 & 32.1 $\pm$ 1.3 & 46.7 $\pm$ 1.2 & 31.6 $\pm$ 1.0 \\
    & 30 & 29.2 $\pm$ 0.8 $\bullet$ & 31.0 $\pm$ 0.9 & 33.9 $\pm$ 1.9 & 49.4 $\pm$ 0.8 & 43.1 $\pm$ 0.8 & 41.3 $\pm$ 1.3 & 44.0 $\pm$ 1.8 & 35.0 $\pm$ 1.3 & 49.4 $\pm$ 0.7 & 33.8 $\pm$ 1.3 \\

    \midrule

    \multirow{3}{*}{Philosophy} & 10 & 25.2 $\pm$ 0.6 $\bullet$ & 27.1 $\pm$ 1.0 & 29.1 $\pm$ 1.0 & 39.8 $\pm$ 1.0 & 36.3 $\pm$ 1.3 & 34.3 $\pm$ 0.9 & 34.4 $\pm$ 1.7 & 30.0 $\pm$ 1.6 & 40.9 $\pm$ 0.9 & 25.5 $\pm$ 1.3 \\
    & 20 & 29.6 $\pm$ 0.4 $\bullet$ & 30.3 $\pm$ 0.9 & 32.3 $\pm$ 0.8 & 43.4 $\pm$ 0.6 & 39.8 $\pm$ 0.9 & 36.0 $\pm$ 1.5 & 36.4 $\pm$ 1.6 & 33.5 $\pm$ 1.3 & 44.0 $\pm$ 0.6 & 30.2 $\pm$ 1.6 \\
    & 30 & 31.8 $\pm$ 0.2 $\bullet$ & 32.3 $\pm$ 1.2 & 35.1 $\pm$ 1.3 & 45.3 $\pm$ 0.8 & 41.8 $\pm$ 0.9 & 39.3 $\pm$ 1.2 & 38.5 $\pm$ 1.7 & 36.0 $\pm$ 1.6 & 45.8 $\pm$ 0.7 & 32.7 $\pm$ 1.6 \\
    
    \bottomrule
\end{tabular}
}
\label{tab:coverage}
\end{table*}

\begin{table*}[!ht]
\renewcommand\arraystretch{1} 
\centering
\caption{Experimental results on \textit{hamming loss} (\%, the smaller the better $\downarrow$). $\bullet$ denotes the best result achieved among all methods.}
\resizebox{\textwidth}{!}{
\begin{tabular}{c c l l l l l l l l l l}
    \toprule
    Data & r & \multicolumn{1}{c}{Schirn} & \multicolumn{1}{c}{NLR} & \multicolumn{1}{c}{FPML} & \multicolumn{1}{c}{PML-LRS} & \multicolumn{1}{c}{PML-NI} & \multicolumn{1}{c}{P-MAP} & \multicolumn{1}{c}{P-VLS} & \multicolumn{1}{c}{PAKS} & \multicolumn{1}{c}{GLC} & \multicolumn{1}{c}{PARD} \\ 
    
    \midrule
     
    Music\_emotion & & 20.2 $\pm$ 0.3 $\bullet$ & 22.2 $\pm$ 0.4 & 23.3 $\pm$ 0.2 & 25.6 $\pm$ 0.5 & 21.6 $\pm$ 0.2 & 22.7 $\pm$ 0.4 & 21.2 $\pm$ 0.5 & 21.5 $\pm$ 0.4 & 21.5 $\pm$ 0.2 & 22.0 $\pm$ 0.2 \\

    Music\_style & & 11.3 $\pm$ 0.2 $\bullet$ & 12.0 $\pm$ 0.3 & 12.4 $\pm$ 0.3 & 16.1 $\pm$ 3.2 & 11.4 $\pm$ 0.2 & 11.7 $\pm$ 0.2 & 11.9 $\pm$ 0.2 & 11.6 $\pm$ 0.2 & 11.5 $\pm$ 0.3 & 14.4 $\pm$ 0.2 \\

    YeastMF & & 9.3 $\pm$ 0.1 & 10.6 $\pm$ 0.7 & 11.7 $\pm$ 0.5 & 11.7 $\pm$ 0.3 & 10.7 $\pm$ 0.2 & 13.3 $\pm$ 0.1 & 11.5 $\pm$ 0.5 & 10.2 $\pm$ 0.2 & 11.3 $\pm$ 0.4 & 9.0 $\pm$ 0.2 $\bullet$ \\

    YeastCC & & 8.3 $\pm$ 0.2 & 8.9 $\pm$ 0.9 & 9.0 $\pm$ 0.3 & 10.8 $\pm$ 0.2 & 9.4 $\pm$ 0.3 & 11.3 $\pm$ 0.4 & 9.5 $\pm$ 0.4 & 7.7 $\pm$ 0.3 $\bullet$ & 10.2 $\pm$ 0.3 & 8.3 $\pm$ 0.2 \\

    YeastBP & & 3.9 $\pm$ 0.1 $\bullet$ & 4.6 $\pm$ 0.2 & 4.3 $\pm$ 0.1 & 5.2 $\pm$ 0.1 & 4.8 $\pm$ 0.0 & 5.3 $\pm$ 0.2 & 4.0 $\pm$ 0.1 & 4.2 $\pm$ 0.1 & 5.1 $\pm$ 0.1 & 3.9 $\pm$ 0.1 $\bullet$ \\
    
    \midrule

    \multirow{3}{*}{Scene} & 1 & 10.8 $\pm$ 0.3 $\bullet$ & 16.3 $\pm$ 2.5 & 20.3 $\pm$ 1.1 & 28.8 $\pm$ 0.4 & 15.2 $\pm$ 0.8 & 11.5 $\pm$ 0.7 & 12.8 $\pm$ 0.7 & 14.4 $\pm$ 1.2 & 15.9 $\pm$ 0.9 & 15.7 $\pm$ 1.1 \\ 

    & 2 & 11.7 $\pm$ 0.8 $\bullet$ & 18.6 $\pm$ 1.3 & 20.7 $\pm$ 0.6 & 29.5 $\pm$ 1.6 & 18.7 $\pm$ 0.9 & 13.1 $\pm$ 0.7 & 14.7 $\pm$ 0.5 & 18.5 $\pm$ 0.9 & 19.5 $\pm$ 0.7 & 17.9 $\pm$ 0.2 \\ 

    & 3 & 16.2 $\pm$ 0.5 $\bullet$ & 21.5 $\pm$ 2.1 & 19.8 $\pm$ 0.9 & 29.3 $\pm$ 1.3 & 23.1 $\pm$ 0.4 & 17.3 $\pm$ 0.8 & 20.4 $\pm$ 0.9 & 23.1 $\pm$ 0.7 & 23.6 $\pm$ 0.5 & 17.9 $\pm$ 0.2 \\ 
    
    \midrule

    \multirow{3}{*}{Birds} & 3 & 8.6 $\pm$ 0.6 $\bullet$ & 10.1 $\pm$ 0.6 & 11.7 $\pm$ 1.7 & 11.5 $\pm$ 1.7 & 10.1 $\pm$ 0.6 & 14.6 $\pm$ 1.9 & 18.8 $\pm$ 2.3 & 11.0 $\pm$ 0.5 & 10.3 $\pm$ 0.6 & 10.9 $\pm$ 2.0 \\ 

    & 7 & 9.5 $\pm$ 0.4 $\bullet$ & 13.3 $\pm$ 1.0 & 12.2 $\pm$ 0.8 & 12.9 $\pm$ 4.5 & 12.9 $\pm$ 0.7 & 17.6 $\pm$ 0.9 & 9.7 $\pm$ 0.7 & 11.4 $\pm$ 0.6 & 13.2 $\pm$ 0.4 & 11.0 $\pm$ 2.4 \\ 

    & 11 & 12.2 $\pm$ 1.0 $\bullet$ & 15.5 $\pm$ 1.4 & 13.4 $\pm$ 2.2 & 15.4 $\pm$ 3.3 & 13.5 $\pm$ 0.4 & 17.2 $\pm$ 0.8 & 13.3 $\pm$ 0.9 & 12.5 $\pm$ 0.7 & 13.5 $\pm$ 0.5 & 13.6 $\pm$ 1.6 \\ 
    
    \midrule

    \multirow{3}{*}{Medical} & 3 & 1.0 $\pm$ 0.1 $\bullet$ & 1.2 $\pm$ 0.2 & 2.1 $\pm$ 0.1 & 3.9 $\pm$ 0.1 & 1.3 $\pm$ 0.1 & 3.0 $\pm$ 0.2 & 2.1 $\pm$ 0.3 & 2.8 $\pm$ 0.1 & 1.9 $\pm$ 0.1 & 2.6 $\pm$ 0.1 \\ 

    & 7 & 1.2 $\pm$ 0.2 $\bullet$ & 1.3 $\pm$ 0.1 $\bullet$ & 2.1 $\pm$ 0.1 & 3.8 $\pm$ 0.1 & 1.7 $\pm$ 0.1 & 3.4 $\pm$ 0.3 & 2.4 $\pm$ 0.1 & 2.9 $\pm$ 0.2 & 3.3 $\pm$ 0.3 & 2.6 $\pm$ 0.1 \\ 

    & 11 & 1.8 $\pm$ 0.1 $\bullet$ & 2.3 $\pm$ 0.4 & 2.1 $\pm$ 0.1 & 3.8 $\pm$ 0.1 & 2.2 $\pm$ 0.3 & 4.5 $\pm$ 0.7 & 2.3 $\pm$ 0.3 & 2.8 $\pm$ 0.2 & 4.3 $\pm$ 0.4 & 2.5 $\pm$ 0.1 \\ 
    
    \midrule

    \multirow{3}{*}{Enron} & 3 & 4.7 $\pm$ 0.2 $\bullet$ & 5.4 $\pm$ 0.5 & 5.7 $\pm$ 0.1 & 11.1 $\pm$ 1.4 & 5.4 $\pm$ 0.2 & 5.5 $\pm$ 0.2 & 6.3 $\pm$ 0.2 & 5.3 $\pm$ 0.1 & 6.7 $\pm$ 0.2 & 5.7 $\pm$ 0.1 \\ 

    & 7 & 4.8 $\pm$ 0.2 $\bullet$ & 5.4 $\pm$ 0.6 & 5.7 $\pm$ 0.1 & 12.0 $\pm$ 1.3 & 6.0 $\pm$ 0.2 & 5.9 $\pm$ 0.3 & 6.4 $\pm$ 0.2 & 5.3 $\pm$ 0.1 & 7.7 $\pm$ 0.2 & 5.7 $\pm$ 0.2 \\ 

    & 11 & 4.9 $\pm$ 0.2 $\bullet$ & 5.6 $\pm$ 0.4 & 5.8 $\pm$ 0.2 & 10.8 $\pm$ 1.6 & 6.7 $\pm$ 0.2 & 5.9 $\pm$ 0.2 & 6.4 $\pm$ 0.2 & 5.3 $\pm$ 0.2 & 8.6 $\pm$ 0.2 & 5.7 $\pm$ 0.2 \\ 
    
    \midrule

    \multirow{3}{*}{Chess} & 10 & 1.0 $\pm$ 0.0 $\bullet$ & 1.1 $\pm$ 0.0 & 1.1 $\pm$ 0.1 & 9.1 $\pm$ 1.7 & 1.4 $\pm$ 0.0 & 1.4 $\pm$ 0.1 & 1.1 $\pm$ 0.0 & 1.1 $\pm$ 0.0 & 1.7 $\pm$ 0.0 & 1.2 $\pm$ 0.0 \\ 

    & 20 & 1.1 $\pm$ 0.1 $\bullet$ & 1.2 $\pm$ 0.0 & 1.1 $\pm$ 0.1 $\bullet$ & 6.9 $\pm$ 1.3 & 1.7 $\pm$ 0.0 & 1.5 $\pm$ 0.0 & 1.1 $\pm$ 0.0 $\bullet$ & 1.2 $\pm$ 0.0 & 2.0 $\pm$ 0.0 & 2.0 $\pm$ 0.1 \\ 

    & 30 & 1.1 $\pm$ 0.1 $\bullet$ & 1.2 $\pm$ 0.0 & 1.2 $\pm$ 0.1 & 4.7 $\pm$ 1.3 & 1.8 $\pm$ 0.0 & 1.6 $\pm$ 0.0 & 1.1 $\pm$ 0.0 $\bullet$ & 1.2 $\pm$ 0.1 & 2.2 $\pm$ 0.0 & 3.6 $\pm$ 0.1 \\ 
    
    \midrule

    \multirow{3}{*}{Philosophy} & 10 & 1.0 $\pm$ 0.0 $\bullet$ & 1.1 $\pm$ 0.0 & 1.0 $\pm$ 0.0 $\bullet$ & 7.2 $\pm$ 0.5 & 1.3 $\pm$ 0.0 & 1.3 $\pm$ 0.0 & 1.0 $\pm$ 0.0 $\bullet$ & 1.0 $\pm$ 0.0 $\bullet$ & 1.5 $\pm$ 0.0 & 1.0 $\pm$ 0.0 $\bullet$ \\ 

    & 20 & 1.0 $\pm$ 0.0 $\bullet$ & 1.1 $\pm$ 0.0 & 1.1 $\pm$ 0.0 & 6.3 $\pm$ 0.5 & 1.6 $\pm$ 0.0 & 1.3 $\pm$ 0.1 & 1.0 $\pm$ 0.0 $\bullet$ & 1.2 $\pm$ 0.0 & 1.8 $\pm$ 0.0 & 1.0 $\pm$ 0.0 $\bullet$ \\ 

    & 30 & 1.0 $\pm$ 0.0 $\bullet$ & 1.1 $\pm$ 0.0 & 1.1 $\pm$ 0.0 & 5.0 $\pm$ 0.4 & 1.8 $\pm$ 0.0 & 1.4 $\pm$ 0.1 & 1.1 $\pm$ 0.0 & 1.3 $\pm$ 0.0 & 2.0 $\pm$ 0.0 & 1.0 $\pm$ 0.0 $\bullet$ \\ 
    
    \bottomrule
\end{tabular}
}
\label{tab:hamming loss}
\end{table*}


\begin{table*}[!ht]
\renewcommand\arraystretch{1} 
\centering
\caption{Experimental results on $\textit{one-error}$ (\%, lower is better $\downarrow$). $\bullet$ denotes the best result achieved among all methods.}
\resizebox{\textwidth}{!}{
\begin{tabular}{c c l l l l l l l l l l}
    \toprule
    Data & r & Schirn & NLR & FPML & PML-LRS & PML-NI & P-MAP & P-VLS & PAKS & GLC & PARD \\
    \midrule

    Music\_emotion & & 45.0 $\pm$ 0.8 & 51.0 $\pm$ 2.1 & 56.1 $\pm$ 0.9 & 45.9 $\pm$ 1.4 & 47.7 $\pm$ 1.6 & 52.4 $\pm$ 2.9 & {36.2} $\pm$ 5.1 $\bullet$ & 46.6 $\pm$ 1.9 & 46.1 $\pm$ 1.6 & 46.7 $\pm$ 1.9 \\
    Music\_style & & {32.6} $\pm$ 0.2 $\bullet$ & 34.4 $\pm$ 0.8 & 39.9 $\pm$ 1.2 & 34.4 $\pm$ 1.5 & 34.3 $\pm$ 1.2 & 36.7 $\pm$ 0.3 & 35.8 $\pm$ 0.8 & 36.1 $\pm$ 0.6 & 34.3 $\pm$ 1.6 & 34.8 $\pm$ 1.6 \\
    YeastMF & & {47.0} $\pm$ 2.2 $\bullet$ & 52.8 $\pm$ 1.9 & 75.9 $\pm$ 3.3 & 74.4 $\pm$ 2.8 & 63.4 $\pm$ 2.7 & 95.2 $\pm$ 1.8 & 85.9 $\pm$ 4.7 & 59.3 $\pm$ 0.9 & 70.5 $\pm$ 3.2 & 73.6 $\pm$ 2.8 \\
    YeastCC & & {31.5} $\pm$ 1.2 $\bullet$ & 34.3 $\pm$ 1.8 & 53.6 $\pm$ 2.4 & 71.7 $\pm$ 1.2 & 54.4 $\pm$ 1.3 & 84.5 $\pm$ 1.8 & 72.2 $\pm$ 2.6 & 36.6 $\pm$ 2.1 & 67.5 $\pm$ 1.0 & 70.7 $\pm$ 3.3 \\
    YeastBP & & {48.6} $\pm$ 0.3 $\bullet$ & 50.7 $\pm$ 0.8 & 63.4 $\pm$ 0.7 & 81.7 $\pm$ 1.0 & 67.2 $\pm$ 1.3 & 82.2 $\pm$ 2.8 & 74.9 $\pm$ 2.6 & 53.1 $\pm$ 1.8 & 78.3 $\pm$ 1.9 & 64.3 $\pm$ 1.6 \\
    
    \midrule

    \multirow{3}{*}{Scene} 
    & 1 & {23.4} $\pm$ 1.1 $\bullet$ & 24.5 $\pm$ 1.8 & 50.8 $\pm$ 4.6 & 38.9 $\pm$ 2.3 & 38.1 $\pm$ 2.3 & 30.0 $\pm$ 2.4 & 25.6 $\pm$ 1.4 & 36.4 $\pm$ 3.4 & 38.9 $\pm$ 2.3 & 40.1 $\pm$ 4.5 \\
    & 2 & {23.9} $\pm$ 2.5 $\bullet$ & 25.8 $\pm$ 2.9 & 50.3 $\pm$ 2.9 & 49.5 $\pm$ 2.7 & 47.7 $\pm$ 2.9 & 32.9 $\pm$ 2.1 & 27.1 $\pm$ 2.6 & 46.9 $\pm$ 3.3 & 49.1 $\pm$ 2.5 & 50.3 $\pm$ 2.8 \\
    & 3 & {26.1} $\pm$ 1.5 $\bullet$ & 28.7 $\pm$ 2.6 & 51.6 $\pm$ 4.2 & 59.9 $\pm$ 1.2 & 58.7 $\pm$ 1.3 & 43.7 $\pm$ 2.1 & 34.3 $\pm$ 0.7 & 59.6 $\pm$ 1.9 & 59.7 $\pm$ 1.3 & 60.6 $\pm$ 1.4 \\
    
    \midrule

    \multirow{3}{*}{Birds} 
    & 3 & {41.0} $\pm$ 1.3 $\bullet$ & 47.6 $\pm$ 1.8 & 71.1 $\pm$ 4.5 & 54.9 $\pm$ 5.7 & 49.7 $\pm$ 7.0 & 69.6 $\pm$ 10.0 & 64.1 $\pm$ 5.9 & 60.2 $\pm$ 5.7 & 52.0 $\pm$ 5.5 & 84.2 $\pm$ 6.0 \\
    & 7 & {50.2} $\pm$ 2.5 $\bullet$ & 54.7 $\pm$ 3.3 & 78.9 $\pm$ 7.2 & 83.6 $\pm$ 4.4 & 76.5 $\pm$ 3.4 & 86.3 $\pm$ 2.8 & 69.9 $\pm$ 5.0 & 60.1 $\pm$ 5.9 & 85.6 $\pm$ 4.9 & 89.1 $\pm$ 0.8 \\
    & 11 & {64.9} $\pm$ 2.8 $\bullet$ & 66.3 $\pm$ 2.8 & 80.0 $\pm$ 7.1 & 86.7 $\pm$ 2.4 & 82.7 $\pm$ 6.8 & 91.3 $\pm$ 5.8 & 80.0 $\pm$ 3.7 & 75.2 $\pm$ 5.4 & 83.9 $\pm$ 5.0 & 91.9 $\pm$ 3.9 \\

    \midrule

    \multirow{3}{*}{Medical} 
    & 3 & {11.5} $\pm$ 1.9 $\bullet$ & 14.6 $\pm$ 3.1 & 20.0 $\pm$ 1.2 & 27.7 $\pm$ 1.5 & 15.3 $\pm$ 1.0 & 40.8 $\pm$ 3.4 & 24.3 $\pm$ 2.5 & 49.9 $\pm$ 1.1 & 25.8 $\pm$ 1.9 & 18.9 $\pm$ 1.8 \\
    & 7 & {14.6} $\pm$ 1.1 $\bullet$ & 21.6 $\pm$ 4.0 & 31.7 $\pm$ 2.5 & 59.2 $\pm$ 3.1 & 20.4 $\pm$ 1.8 & 43.7 $\pm$ 3.6 & 30.5 $\pm$ 4.5 & 50.3 $\pm$ 2.1 & 41.6 $\pm$ 1.9 & 22.4 $\pm$ 3.3 \\
    & 11 & {14.8} $\pm$ 1.8 $\bullet$ & 27.0 $\pm$ 4.6 & 32.2 $\pm$ 2.3 & 72.9 $\pm$ 4.3 & 27.6 $\pm$ 2.8 & 44.6 $\pm$ 6.9 & 30.3 $\pm$ 3.0 & 50.3 $\pm$ 2.3 & 54.0 $\pm$ 5.3 & 27.4 $\pm$ 2.6 \\

    \midrule

    \multirow{3}{*}{Enron} 
    & 3 & {21.2} $\pm$ 1.8 $\bullet$ & 24.0 $\pm$ 2.2 & 31.8 $\pm$ 2.8 & 50.6 $\pm$ 2.6 & 29.7 $\pm$ 2.4 & 29.1 $\pm$ 1.2 & 42.4 $\pm$ 7.0 & 23.8 $\pm$ 2.5 & 45.1 $\pm$ 1.1 & 24.2 $\pm$ 1.6 \\
    & 7 & {20.1} $\pm$ 1.9 $\bullet$ & 28.6 $\pm$ 4.2 & 30.8 $\pm$ 2.5 & 68.7 $\pm$ 1.3 & 35.8 $\pm$ 1.4 & 31.1 $\pm$ 6.2 & 50.5 $\pm$ 10.2 & 24.1 $\pm$ 2.5 & 57.9 $\pm$ 2.0 & 25.7 $\pm$ 1.4 \\
    & 11 & {22.3} $\pm$ 1.9 $\bullet$ & 32.3 $\pm$ 6.5 & 30.6 $\pm$ 2.3 & 76.9 $\pm$ 2.2 & 44.3 $\pm$ 1.9 & 34.8 $\pm$ 3.2 & 56.1 $\pm$ 11.1 & 24.5 $\pm$ 2.1 & 67.0 $\pm$ 0.9 & 27.3 $\pm$ 2.5 \\

    \midrule

    \multirow{3}{*}{Chess} 
    & 10 & {43.1} $\pm$ 1.6 $\bullet$ & 47.1 $\pm$ 4.2 & 47.7 $\pm$ 1.2 & 63.6 $\pm$ 1.6 & 56.1 $\pm$ 1.5 & 69.3 $\pm$ 2.8 & 69.8 $\pm$ 5.3 & 48.1 $\pm$ 1.5 & 65.9 $\pm$ 1.8 & 45.7 $\pm$ 0.7 \\
    & 20 & {44.7} $\pm$ 1.7 $\bullet$ & 47.8 $\pm$ 3.1 & 48.4 $\pm$ 1.7 & 76.9 $\pm$ 1.3 & 62.4 $\pm$ 2.0 & 69.5 $\pm$ 1.4 & 74.3 $\pm$ 6.8 & 49.5 $\pm$ 0.7 & 75.0 $\pm$ 2.1 & 47.7 $\pm$ 0.6 \\
    & 30 & {45.6} $\pm$ 1.7 $\bullet$ & 52.2 $\pm$ 2.5 & 50.5 $\pm$ 2.8 & 84.9 $\pm$ 0.7 & 69.9 $\pm$ 2.1 & 68.5 $\pm$ 1.1 & 84.7 $\pm$ 4.8 & 48.7 $\pm$ 1.7 & 83.0 $\pm$ 0.9 & 51.7 $\pm$ 2.0 \\

    \midrule

    \multirow{3}{*}{Philosophy} 
    & 10 & {46.2} $\pm$ 1.7 $\bullet$ & 50.5 $\pm$ 3.5 & 51.5 $\pm$ 1.2 & 60.7 $\pm$ 2.0 & 56.2 $\pm$ 1.1 & 74.4 $\pm$ 1.3 & 71.0 $\pm$ 1.8 & 47.7 $\pm$ 0.5 & 60.5 $\pm$ 2.0 & 49.3 $\pm$ 1.3 \\
    & 20 & {47.1} $\pm$ 1.3 $\bullet$ & 54.5 $\pm$ 3.2 & 52.3 $\pm$ 1.1 & 70.6 $\pm$ 2.2 & 63.9 $\pm$ 1.7 & 75.6 $\pm$ 1.8 & 66.5 $\pm$ 2.2 & 49.5 $\pm$ 0.5 & 70.3 $\pm$ 2.0 & 52.0 $\pm$ 1.4 \\
    & 30 & {48.9} $\pm$ 1.2 $\bullet$ & 60.5 $\pm$ 2.8 & 53.2 $\pm$ 1.0 & 76.7 $\pm$ 1.9 & 70.3 $\pm$ 1.5 & 76.9 $\pm$ 2.6 & 68.2 $\pm$ 1.6 & 53.3 $\pm$ 0.7 & 76.4 $\pm$ 2.1 & 54.6 $\pm$ 1.7 \\

    \bottomrule
\end{tabular}
}
\label{tab:one error}
\end{table*}


\begin{table*}[!ht]
\renewcommand\arraystretch{1} 
\centering
\caption{Ablation study of Schirn in terms of \textit{coverage} and \textit{hamming loss} on synthetic datasets (in \%). The settings for each synthetic dataset are $r=1$ for Scene, $r=3$ for Birds, Medical, and Enron, and $r=10$ for Chess and Philosophy.}
\resizebox{\textwidth}{!}{
\begin{tabular}{ c c c | c c c c c c | c c c c c c  }
    \toprule
    
    High & \multirow{2}{*}{Sparsity} & Low & \multicolumn{6}{c|}{\textbf{coverage $\downarrow$}} & \multicolumn{6}{c}{\textbf{hamming loss $\downarrow$}} \\ 
    Rank & & Rank & Scene & Birds & Medical & Enron & Chess & Philosophy & Scene & Birds & Medical & Enron & Chess & Philosophy \\

    \midrule

    \ding{53} & \checkmark & \ding{53} & 9.7 $\pm$ 0.8 & 32.4 $\pm$ 2.2 & 5.1 $\pm$ 1.1 & 27.7 $\pm$ 1.2 & 26.4 $\pm$ 1.3 & 27.5 $\pm$ 0.9 & 11.2 $\pm$ 0.2 & 9.2 $\pm$ 0.5 & 1.1 $\pm$ 0.1 & 4.9 $\pm$ 0.2 & 1.1 $\pm$ 0.0 & 1.0 $\pm$ 0.0 \\

    \checkmark & \ding{53} & \ding{53} & 29.2 $\pm$ 2.2 & 46.3 $\pm$ 3.3 & 8.6 $\pm$ 1.3 & 60.5 $\pm$ 2.0 & 63.9 $\pm$ 5.4 & 62.4 $\pm$ 2.5 & 16.8 $\pm$ 0.3 & 9.5 $\pm$ 0.6 & 1.1 $\pm$ 0.1 & 5.2 $\pm$ 0.1 & 1.1 $\pm$ 0.0 & 1.0 $\pm$ 0.0 \\

    \ding{53} & \checkmark &  \checkmark & 9.0 $\pm$ 0.7 & 34.7 $\pm$ 2.8 & 5.1 $\pm$ 1.0 & 27.8 $\pm$ 1.4 & 26.2 $\pm$ 1.8 & 27.3 $\pm$ 0.3 & 11.3 $\pm$ 0.5 & 9.7 $\pm$ 0.5 & 1.1 $\pm$ 0.2 & 5.0 $\pm$ 0.2 & 1.1 $\pm$ 0.0 & 1.0 $\pm$ 0.0 \\

    \midrule
    
    \checkmark & \checkmark & \ding{53} & 7.7 $\pm$ 0.3 & 26.1 $\pm$ 0.5 & 3.5 $\pm$ 0.8 & 24.5 $\pm$ 0.6 & 24.3 $\pm$ 0.9 & 25.2 $\pm$ 0.6 & 10.8 $\pm$ 0.3 & 8.6 $\pm$ 0.6 & 1.0 $\pm$ 0.1 & 4.7 $\pm$ 0.2 & 1.0 $\pm$ 0.0 & 1.0 $\pm$ 0.0 \\

    \bottomrule
\end{tabular}
}       
\label{tab:ablation1}
\end{table*}

\begin{table*}[!ht]\scriptsize
\renewcommand\arraystretch{1} 
    \centering
    \caption{Ablation study of Schirn in terms of \textit{one-error} on synthetic data sets. The settings for each synthetic dataset are $r=1$ for Scene, $r=3$ for Birds, Medical and Enron, and $r=10$ for Chess and Philosophy.}
    \begin{tabular}{ c c c | c c c c c c  }
        \toprule
        
        High & \multirow{2}{*}{Sparsity} & Low & \multicolumn{6}{c}{\textbf{one-error $\downarrow$}} \\ 
        Rank & & Rank & Scene & Birds & Medical & Enron & Chess & Philosophy \\

        \midrule

        \ding{53} & \checkmark & \ding{53} & .282 $\pm$ .014 & .508 $\pm$ .071 & .138 $\pm$ .022 & .226 $\pm$ .022 & .463 $\pm$ .015 & .484 $\pm$ .006  \\

        \checkmark & \ding{53} & \ding{53} & .602 $\pm$ .020 & .550 $\pm$ .033 & .164 $\pm$ .025 & .326 $\pm$ .021 & .992 $\pm$ .007 & .840 $\pm$ .023  \\

        \ding{53} & \checkmark &  \checkmark & .278 $\pm$ .018 & .502 $\pm$ .070 & .130 $\pm$ .026 & .238 $\pm$ .017 & .468 $\pm$ .032 & .474 $\pm$ .014 \\

        \midrule
        
         \checkmark &  \checkmark & \ding{53} & .234 $\pm$ .011 & .410 $\pm$ .013 & .115 $\pm$ .019 & .212 $\pm$ .018 & .431 $\pm$ .016 & .462 $\pm$ .017 \\

        \bottomrule
    \end{tabular}
    \label{tab:ablation2}
\end{table*}

%% file: main.bbl
\begin{thebibliography}{10}

\bibitem{beck2009fast}
Amir Beck and Marc Teboulle.
\newblock A fast iterative shrinkage-thresholding algorithm for linear inverse problems.
\newblock {\em SIAM journal on imaging sciences}, 2(1):183--202, 2009.

\bibitem{BOUTELL2004Learning}
Matthew~R Boutell, Jiebo Luo, Xipeng Shen, and Christopher~M Brown.
\newblock Learning multi-label scene classification.
\newblock {\em Pattern recognition}, 37(9):1757--1771, 2004.

\bibitem{cai2010singular}
Jian-Feng Cai, Emmanuel~J Cand{\`e}s, and Zuowei Shen.
\newblock A singular value thresholding algorithm for matrix completion.
\newblock {\em SIAM Journal on optimization}, 20(4):1956--1982, 2010.

\bibitem{candes2012exact}
Emmanuel Candes and Benjamin Recht.
\newblock Exact matrix completion via convex optimization.
\newblock {\em Communications of the ACM}, 55(6):111--119, 2012.

\bibitem{candes2005decoding}
Emmanuel~J Candes and Terence Tao.
\newblock Decoding by linear programming.
\newblock {\em IEEE transactions on information theory}, 51(12):4203--4215, 2005.

\bibitem{cheng2017iatc}
Xiang Cheng, Shu-Guang Zhao, Xuan Xiao, and Kuo-Chen Chou.
\newblock iatc-misf: a multi-label classifier for predicting the classes of anatomical therapeutic chemicals.
\newblock {\em Bioinformatics}, 33(3):341--346, 2017.

\bibitem{Johannes2008Multilabel}
J~Fürnkranz, E~Hüllermeier, E~Loza?Mencía, and K.~Brinker.
\newblock Multilabel classification via calibrated label ranking.
\newblock {\em Machine Learning}, 73(2):133--153, 2008.

\bibitem{gibaja2015tutorial}
Eva Gibaja and Sebasti{\'a}n Ventura.
\newblock A tutorial on multilabel learning.
\newblock {\em ACM Computing Surveys (CSUR)}, 47(3):1--38, 2015.

\bibitem{hang2023partial}
Jun-Yi Hang and Min-Ling Zhang.
\newblock Partial multi-label learning with probabilistic graphical disambiguation.
\newblock {\em Advances in Neural Information Processing Systems}, 36:1339--1351, 2023.

\bibitem{huang2015learning}
Jun Huang, Guorong Li, Qingming Huang, and Xindong Wu.
\newblock Learning label specific features for multi-label classification.
\newblock In {\em 2015 IEEE International conference on data mining}, pages 181--190. IEEE, 2015.

\bibitem{kuznetsova2020open}
Alina Kuznetsova, Hassan Rom, Neil Alldrin, Jasper Uijlings, Ivan Krasin, Jordi Pont-Tuset, Shahab Kamali, Stefan Popov, Matteo Malloci, Alexander Kolesnikov, et~al.
\newblock The open images dataset v4: Unified image classification, object detection, and visual relationship detection at scale.
\newblock {\em International journal of computer vision}, 128(7):1956--1981, 2020.

\bibitem{lai2016instance}
Hanjiang Lai, Pan Yan, Xiangbo Shu, Yunchao Wei, and Shuicheng Yan.
\newblock Instance-aware hashing for multi-label image retrieval.
\newblock {\em IEEE Transactions on Image Processing}, 25(6):2469--2479, 2016.

\bibitem{lin2010augmented}
Zhouchen Lin, Minming Chen, and Yi~Ma.
\newblock The augmented lagrange multiplier method for exact recovery of corrupted low-rank matrices.
\newblock {\em arXiv preprint arXiv:1009.5055}, 2010.

\bibitem{lyu2021prior}
Gengyu Lyu, Songhe Feng, Yi~Jin, Tao Wang, Congyan Lang, and Yidong Li.
\newblock Prior knowledge regularized self-representation model for partial multilabel learning.
\newblock {\em IEEE Transactions on Cybernetics}, 53(3):1618--1628, 2021.

\bibitem{lyu2020partial}
Gengyu Lyu, Songhe Feng, and Yidong Li.
\newblock Partial multi-label learning via probabilistic graph matching mechanism.
\newblock In {\em Proceedings of the 26th ACM SIGKDD International Conference on Knowledge Discovery \& Data Mining}, pages 105--113, 2020.

\bibitem{ma2021expand}
Zhongchen Ma and Songcan Chen.
\newblock Expand globally, shrink locally: Discriminant multi-label learning with missing labels.
\newblock {\em Pattern Recognition}, 111:107675, 2021.

\bibitem{qi2007correlative}
Guo-Jun Qi, Xian-Sheng Hua, Yong Rui, Jinhui Tang, Tao Mei, and Hong-Jiang Zhang.
\newblock Correlative multi-label video annotation.
\newblock In {\em Proceedings of the 15th ACM international conference on Multimedia}, pages 17--26, 2007.

\bibitem{2009Classifier1212}
J.~Read, B.~Pfahringer, G.~Holmes, and E.~Frank.
\newblock Classifier chains for multi-label classification.
\newblock In {\em Proceedings of the European Conference on Machine Learning and Knowledge Discovery in Databases: Part II}, 2009.

\bibitem{rockafellar1974augmented}
R~Tyrrell Rockafellar.
\newblock Augmented lagrange multiplier functions and duality in nonconvex programming.
\newblock {\em SIAM Journal on Control}, 12(2):268--285, 1974.

\bibitem{si2023multi}
Chongjie Si, Yuheng Jia, Ran Wang, Min-Ling Zhang, Yanghe Feng, and Chongxiao Qu.
\newblock Multi-label classification with high-rank and high-order label correlations.
\newblock {\em IEEE Transactions on Knowledge and Data Engineering}, 36(8):4076--4088, 2023.

\bibitem{sun2021global}
Lijuan Sun, Songhe Feng, Jun Liu, Gengyu Lyu, and Congyan Lang.
\newblock Global-local label correlation for partial multi-label learning.
\newblock {\em IEEE Transactions on Multimedia}, 24:581--593, 2021.

\bibitem{sun2021partial}
Lijuan Sun, Songhe Feng, Gengyu Lyu, Hua Zhang, and Guojun Dai.
\newblock Partial multi-label learning with noisy side information.
\newblock {\em Knowledge and Information Systems}, 63:541--564, 2021.

\bibitem{sun2019partial}
Lijuan Sun, Songhe Feng, Tao Wang, Congyan Lang, and Yi~Jin.
\newblock Partial multi-label learning by low-rank and sparse decomposition.
\newblock In {\em Proceedings of the AAAI conference on artificial intelligence}, volume~33, pages 5016--5023, 2019.

\bibitem{tang2009large}
Lei Tang, Suju Rajan, and Vijay~K Narayanan.
\newblock Large scale multi-label classification via metalabeler.
\newblock In {\em Proceedings of the 18th international conference on World wide web}, pages 211--220, 2009.

\bibitem{tsoumakas2010random}
Grigorios Tsoumakas, Ioannis Katakis, and Ioannis Vlahavas.
\newblock Random k-labelsets for multilabel classification.
\newblock {\em IEEE Transactions on Knowledge and Data Engineering}, 23(7):1079--1089, 2011.

\bibitem{wang2019discriminative}
Haobo Wang, Weiwei Liu, Yang Zhao, Chen Zhang, Tianlei Hu, and Gang Chen.
\newblock Discriminative and correlative partial multi-label learning.
\newblock In {\em IJCAI}, pages 3691--3697, 2019.

\bibitem{wang2023deep}
Haobo Wang, Shisong Yang, Gengyu Lyu, Weiwei Liu, Tianlei Hu, Ke~Chen, Songhe Feng, and Gang Chen.
\newblock Deep partial multi-label learning with graph disambiguation.
\newblock {\em arXiv preprint arXiv:2305.05882}, 2023.

\bibitem{wang2022partial}
Jing Wang, Peipei Li, and Kui Yu.
\newblock Partial multi-label feature selection.
\newblock In {\em 2022 International Joint Conference on Neural Networks (IJCNN)}, pages 1--9. IEEE, 2022.

\bibitem{9084698}
Xiaoying Wang, Jun Xie, Lu~Yu, and Xingliu Tao.
\newblock Ml-lrc: Low-rank-constraint-based multi-label learning with label noise.
\newblock In {\em 2020 IEEE 4th Information Technology, Networking, Electronic and Automation Control Conference (ITNEC)}, volume~1, pages 129--136, 2020.

\bibitem{wedin1972perturbation}
Per-{\AA}ke Wedin.
\newblock Perturbation bounds in connection with singular value decomposition.
\newblock {\em BIT Numerical Mathematics}, 12:99--111, 1972.

\bibitem{xie2018partial}
Ming-Kun Xie and Sheng-Jun Huang.
\newblock Partial multi-label learning.
\newblock In {\em Proceedings of the AAAI conference on artificial intelligence}, volume~32, 2018.

\bibitem{xie2021partial}
Ming-Kun Xie and Sheng-Jun Huang.
\newblock Partial multi-label learning with noisy label identification.
\newblock {\em IEEE Transactions on Pattern Analysis and Machine Intelligence}, 44(7):3676--3687, 2021.

\bibitem{xu2020partial}
Ning Xu, Yun-Peng Liu, and Xin Geng.
\newblock Partial multi-label learning with label distribution.
\newblock In {\em Proceedings of the AAAI conference on artificial intelligence}, volume~34, pages 6510--6517, 2020.

\bibitem{yang2024noisy}
Fuchao Yang, Yuheng Jia, Hui Liu, Yongqiang Dong, and Junhui Hou.
\newblock Noisy label removal for partial multi-label learning.
\newblock In {\em Proceedings of the 30th ACM SIGKDD Conference on Knowledge Discovery and Data Mining}, pages 3724--3735, 2024.

\bibitem{yu2018feature}
Guoxian Yu, Xia Chen, Carlotta Domeniconi, Jun Wang, Zhao Li, Zili Zhang, and Xindong Wu.
\newblock Feature-induced partial multi-label learning.
\newblock In {\em 2018 IEEE international conference on data mining (ICDM)}, pages 1398--1403. IEEE, 2018.

\bibitem{yu2020partial}
Tingting Yu, Guoxian Yu, Jun Wang, Carlotta Domeniconi, and Xiangliang Zhang.
\newblock Partial multi-label learning using label compression.
\newblock In {\em 2020 IEEE International Conference on Data Mining (ICDM)}, pages 761--770. IEEE, 2020.

\bibitem{zhang2020partial}
Min-Ling Zhang and Jun-Peng Fang.
\newblock Partial multi-label learning via credible label elicitation.
\newblock {\em IEEE Transactions on Pattern Analysis and Machine Intelligence}, 43(10):3587--3599, 2020.

\bibitem{zhang2018binary}
Min-Ling Zhang, Yu-Kun Li, Xu-Ying Liu, and Xin Geng.
\newblock Binary relevance for multi-label learning: an overview.
\newblock {\em Frontiers of Computer Science}, 12:191--202, 2018.

\bibitem{zhang2014lift}
Min-Ling Zhang and Lei Wu.
\newblock Lift: Multi-label learning with label-specific features.
\newblock {\em IEEE transactions on pattern analysis and machine intelligence}, 37(1):107--120, 2014.

\bibitem{zhang2006multilabel}
Min-Ling Zhang and Zhi-Hua Zhou.
\newblock Multilabel neural networks with applications to functional genomics and text categorization.
\newblock {\em IEEE transactions on Knowledge and Data Engineering}, 18(10):1338--1351, 2006.

\bibitem{zhang2007mlds}
Min-Ling Zhang and Zhi-Hua Zhou.
\newblock Ml-knn: A lazy learning approach to multi-label learning.
\newblock {\em Pattern recognition}, 40(7):2038--2048, 2007.

\bibitem{zhang2013review}
Min-Ling Zhang and Zhi-Hua Zhou.
\newblock A review on multi-label learning algorithms.
\newblock {\em IEEE transactions on knowledge and data engineering}, 26(8):1819--1837, 2013.

\bibitem{zhao2022partial}
Peng Zhao, Shiyi Zhao, Xuyang Zhao, Huiting Liu, and Xia Ji.
\newblock Partial multi-label learning based on sparse asymmetric label correlations.
\newblock {\em Knowledge-Based Systems}, 245:108601, 2022.

\bibitem{2017Multisdfsaf}
Yue Zhu, James~T. Kwok, and Zhi-Hua Zhou.
\newblock Multi-label learning with global and local label correlation.
\newblock {\em IEEE Transactions on Knowledge and Data Engineering}, 30(6):1081--1094, 2018.

\bibitem{zhuang2012non}
Liansheng Zhuang, Haoyuan Gao, Zhouchen Lin, Yi~Ma, Xin Zhang, and Nenghai Yu.
\newblock Non-negative low rank and sparse graph for semi-supervised learning.
\newblock In {\em 2012 IEEE Conference on Computer Vision and Pattern Recognition}, pages 2328--2335. IEEE, 2012.

\end{thebibliography}
